\documentclass{ws-procs11x85}
\usepackage{ws-procs-thm}

\usepackage[main=english]{babel} % hyphenation
\usepackage{microtype} % typography (kerning)
\usepackage[babel,autostyle]{csquotes} % proper quotes
\usepackage{booktabs,multirow}
\usepackage{placeins}

\usepackage{hyperref}
\hypersetup{
    colorlinks=false,
    pdftitle={Coreopsis},
    pdfpagemode=FullScreen,
    }

\usepackage{graphicx}
\graphicspath{ {./img/} }

\begin{document}

\title{Federated generative event models for tokenized electronic health records}

\author{Michael C. Burkhart, Luke Solo, Inhyeok Lee, S'Khaja Charles, \\Zewei ``Whiskey'' Liao, Kaveri Chhikara, Dema Therese}
\address{Department of Medicine, University of Chicago,\\
	Chicago, Illinois, USA\\
	E-mail: \{burkh4rt,lsolo,ihlee,skhaja,wliao0504,kaveri,dema\}@uchicago.edu}

\author{Wan-Ting Liao, Catherine A. Gao}
\address{Department of Medicine, Northwestern University Feinberg School of Medicine,\\
	Chicago, Illinois, USA\\
	E-mail: \{wanting.liao,catherine.gao\}@northwestern.edu}

\author{William F. Parker and Brett K. Beaulieu-Jones}
\address{Department of Medicine, University of Chicago,\\
	Chicago, Illinois, USA\\
	E-mail: \{wparker,beaulieujones\}@uchicago.edu}

\begin{abstract}
Electronic health record foundation models are limited by institutionally siloed data and substantial performance degradation under cross-site transfer. We evaluated federated training of tokenized generative event models (GEMs) across 122,251 intensive care hospitalizations from three independent health systems harmonized to the Common Longitudinal ICU Data Format. Models were assessed on 12 post-24-hour clinical prediction tasks using within-site, cross-site, centralized, and federated training configurations. GEMs achieved the highest mean within-site and cross-site ROC-AUC and were substantially more transportable than conventional supervised models: their average cross-site penalties were 0.025 ROC-AUC and 0.027 PR-AUC, compared with 0.079 and 0.089 for LightGBM. Federated Learning (FedAvg and FedAvgM) approached the performance of centralized GEM training, with most gains obtained within 5--10 communication rounds. However, centralized multi-site training provided only modest improvements over complete local training. Multi-site models were most useful when local training data were limited, with their advantage narrowing as institutional data accumulated. These findings show that federated GEM training is technically feasible and preserves most centralized performance, but that the main open challenge is learning transportable representations to translate larger, but heterogeneous data from multiple health systems into a reliable target-site benefit.
\end{abstract}


\copyrightinfo{\copyright\ 2026 The Authors.}
\clearpage

\section{Introduction}
Transformer models for language benefit from massive collections of publicly available training data~\cite{Kap20,Hof22}. Electronic health records (EHRs), in contrast, are inherently private, institutionally governed, and recorded using site-specific conventions. Consequently, clinical generative event models (GEMs) face two related barriers to scale. First, the amount of data available within any single institution limits model size and performance under established scaling relationships~\cite{Zha25,Wax25}. Second, models trained at one institution frequently lose performance when transferred to another without additional training~\cite{Guo24,Bur25,Pan25}. Scaling clinical GEMs consequently requires not only access to more data, but also methods that can learn across heterogeneous health systems without assuming that patient-level records can be freely pooled.

These two barriers require complementary solutions. Common data models can establish shared clinical semantics across institutions, either through manually curated mappings~\cite{Mur10,Vos15} or through learned cross-institutional correspondences~\cite{Hon21,Zho26}. However, automated correspondence methods remain complex and can perform substantially worse than mappings curated by domain experts. Even after semantic harmonization, governance, privacy, and security concerns may prevent patient-level data from being centralized. These concerns are increasingly salient given the growing frequency and scale of breaches involving protected health information~\cite{Nep22,Jia25}. Federated learning (FL) provides a framework for jointly training models while allowing each participating institution to retain control of its underlying records.

In this work, we study the intersection of these two approaches using ICU data from three health systems: the University of Chicago Medical Center, Northwestern Medicine, and Beth Israel Deaconess Medical Center. The datasets have been mapped by domain experts to the Common Longitudinal ICU Data Format (CLIF)~\cite{Roj25,Lyo26}, providing a shared representation of critical care data across sites. We then evaluate whether federated optimization can recover the benefits of multi-site GEM training without requiring the underlying patient-level datasets to be stored together. This design allows us to distinguish limitations arising from semantic and distributional differences across institutions from limitations introduced by federated learning.

In a standard GEM workflow~\cite{Bur26,Cha26}, each hospitalization is represented as a chronologically ordered sequence of tokens, with each token corresponding to an event in the patient record. For example, \texttt{LAB-ORD//albumin} represents an albumin laboratory order, while \texttt{XFR-OUT//ed} represents a transfer out of the emergency department. The GEM is pretrained from a random initialization to predict the next event token conditioned on the preceding sequence. Once trained, the model can support downstream prediction through representation-based, generative, or supervised fine-tuning approaches~\cite{Guo26b}.

We focus on representation-based inference~\cite{Wor23,Wor25,Lee26,Guo26,Zha26} as opposed to generative inference or supervised finetuning~\cite{McD23,Kra24,Ren24,Ren25,Wax25,Sol26,Ste24,Fal25,Bur25,Cha26}. For each hospitalization, the pretrained GEM produces a fixed-length representation of the observed event sequence. These representations are then used as features for lightweight, outcome-specific classifiers. This approach directly tests whether GEM pretraining produces reusable patient representations and permits controlled comparisons among locally trained, transferred, federated, and centralized models using identical downstream estimators.

We evaluate GEMs in the three independent health systems covering 122,251 ICU hospitalizations and 12 clinical outcomes. Our primary contributions are as follows:

\begin{enumerate}
    \item \textbf{GEM representations are substantially more portable across health systems than representations derived from conventional supervised models.}
    A GEM achieved the highest mean within-site and cross-site ROC-AUCs, while incurring a cross-site penalty of 0.025 ROC-AUC and 0.027 PR-AUC. The corresponding penalties were 0.079 and 0.089 for LightGBM. A one-epoch GEM exhibited the same qualitative pattern, indicating that improved transportability was not restricted to the extended-training configuration.

    \item \textbf{Federation recovers most of the performance of centralized multi-site training, but its practical value is concentrated at data-limited sites.}
    FedAvg and FedAvgM closely approach models trained on centrally pooled data, whereas FedAdam performs substantially worse. Performance gains largely saturate after 5--10 federated rounds, limiting the amount of cross-site communication required. Federated models provide their clearest advantage for newly participating or data-limited hospitals; locally trained models become competitive after approximately 3,000 training examples.

    \item \textbf{The principal limitation in this setting appears to be cross-site transfer and adaptation rather than federated optimization.}
    In our empirical analyses, federated models closely track centralized multi-site models, indicating that the use of federated rather than centralized optimization accounts for only a modest portion of the remaining performance gap. At the same time, centralized multi-site training provides only limited improvements over strong local models at institutions with large training datasets. Together, these findings suggest that the primary challenge is not simply how model updates are aggregated, but how effectively additional heterogeneous data can improve performance at a particular target institution.
\end{enumerate}

We also release \href{https://github.com/bbj-lab/coreopsis}{\texttt{coreopsis}}, an open-source framework for federated GEM training that integrates CLIF-compatible tokenization, local model training, and federated aggregation. Collectively, our results characterize when federated GEM training is likely to be useful and support a practical deployment strategy in which data-limited hospitals begin with a multi-site federated model and increasingly rely on local training or adaptation as institutional data accumulate.

\section{Related work}

Most published studied on FL in healthcare are single-institution simulations~\cite{Teo24,San26}. Many FL methods operate on previously harmonized datasets, such as MIMIC~\cite{Joh23} or eICU~\cite{Pol18}. Dang et al.~\cite{Dan22} benchmarked FL for ICU-related outcomes on eICU and found FedAvg~\cite{McM17} and FedAvgM~\cite{Hsu19} to be strong performers. In MIMIC, Shoham et al.~\cite{Sho24} experimented with FL for diagnosis code prediction and Renc et al.~\cite{Ren25b} proposed federated learning of GEMs for synthetic data generation. 

Generally, cross-site learning relies on a common data model, such as OMOP~\cite{Ove12} or i2b2~\cite{Mur10}. Guo et al.~\cite{Guo24} transferred models trained at Stanford Medicine to Beth Israel and Toronto's Hospital for Sick Children and showed that continued pretraining on local data could quickly match optimal local performance.

Efforts have been made to automatically harmonize datasets. Hong et al.~\cite{Hon21} used embeddings to develop a cross-institutional clinical correspondence. Zhou et al.~\cite{Zho26} learned a correspondence with graph neural networks. Recent efforts have been made to overcome the challenge of data harmonization using language-based representations of events~\cite{Kim26,Guo26c}.

% Representation-based, or rep-based, inference uses the trained model to embed each variable-length sequence of tokens into a fixed-dimensional vector space. These representations are then passed as features to standard off-the-shelf classification models.

\section{Methods}
\subsection{Data}
\label{ss:data}
We considered adult patients (age 18 or older) admitted to the University of Chicago Medical Center between 2018--2024 (UCMC), the Northwestern Medicine healthcare system between 2018--2024 (NU), and Beth Israel Deaconess Medical Center between 2008--2022 (MIMIC-IV-3.1)~\cite{Joh23}. The study was approved by the relevant institutional review boards (IRBs): University of Chicago (IRB20-1823), Northwestern University (STU00202840). MIMIC data were available via credentialed access under data use agreement. For each patient, we considered their first hospitalization and filtered to those that were greater than 24 hours in duration and that involved a transfer to the intensive care unit (ICU) within 24 hours of admission. This yielded 22,000 hospitalizations in UCMC, 65,758 hospitalizations in NU, and 34,495 hospitalizations in MIMIC, each corresponding to unique patients (Table~\ref{tbl:demog}). Each dataset was partitioned into training, tuning, and held-out subsets at a 70\%-10\%-20\% rate. For UCMC and NU, hospitalizations were partitioned according to admission time, with earlier admissions being placed into the training, followed by the tuning, and then the held-out set. For MIMIC, patient privacy is protected by random time shifts at the patient level, making the split into training, tuning, and held-out sets uniformly random.

\begin{table}[tb]
	\tbl{\textbf{Summary statistics of the patients in each dataset}: count and age at admission (mean, standard deviation) are followed percentage breakdowns by  sex, ethnicity, and race.}
	{
		\begin{tabular}{lrrrrrrrrrrr}
			\toprule
			        &         &                        & \multicolumn{1}{c}{sex} & \multicolumn{1}{c}{ethnicity} & \multicolumn{4}{c}{race}                                 \\
			\cmidrule(lr){4-4} \cmidrule(lr){5-5} \cmidrule(lr){6-9} \cmidrule(lr){10-12}
			dataset & count   & age ($\mu \pm \sigma$) & female                  & Hispanic                      & white                    & African Amer. & Asian & other \\              \midrule
			UCMC    & 22,000  & 56.8  $\pm$ 18.3       & 0.41                    & 0.07                          & 0.31                     & 0.55          & 0.02  & 0.13  \\
			NU      & 65,756  & 63.2 $\pm$ 17.3        & 0.44                    & 0.09                          & 0.74                     & 0.12          & 0.04  & 0.10  \\ 
            MIMIC   & 34,495  & 62.7 $\pm$ 18.0        & 0.42                    & 0.03                          & 0.65                     & 0.07          & 0.03  & 0.25  \\ \midrule
			all     & 122,251 & 61.9 $\pm$ 17.9        & 0.43                    & 0.07                          & 0.64                     & 0.18          & 0.03  & 0.15  \\
			\bottomrule
		\end{tabular}
	}
	\label{tbl:demog}
\end{table}

\begin{figure}[tb]
	\centering
    \includegraphics[width=\textwidth]{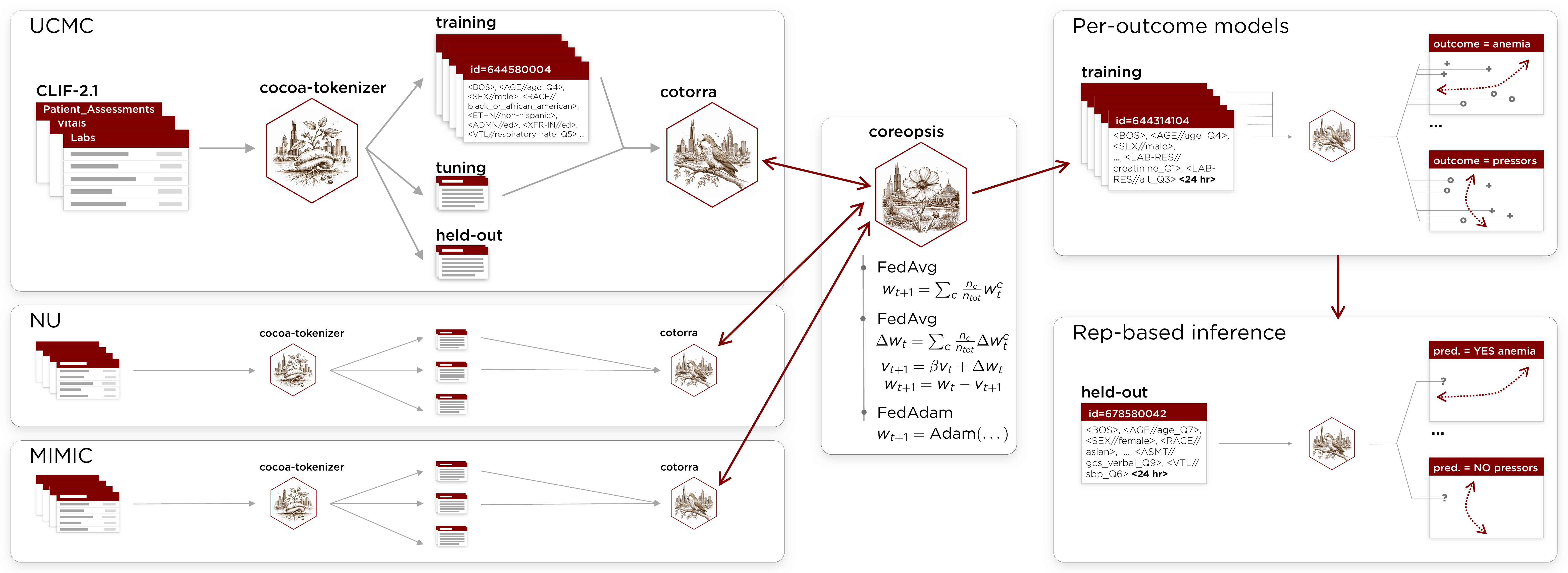}
		\caption{\textbf{Experimental design}: EHR data is converted to the CLIF-2.1 standard in parallel for each of the three sites. A \texttt{cocoa-tokenizer} instance is learned on MIMIC training data and then applied to the other datasets to convert patient records into sequences of tokens. Training and tuning sequences are used at each respective site by a \texttt{cotorra} instance to train a GEM from scratch. The \texttt{coreopsis} software introduced in this paper interacts with each \texttt{cotorra} model to build federated GEMs. To make predictions, per-outcome classifiers are learned on each respective training set using representations extracted from a given model. These models are then applied to representations from the held-out set to perform inference.}
		\label{fig:federation}
\end{figure}

\subsection{Tokenization}
Tokenization proceeded as follows: UCMC, NU, and MIMIC data were converted to the Common Longitudinal ICU Format (CLIF)\cite{Roj25} version 2.1 standard which harmonizes critical care data across institutions by mapping site-specific laboratory and vital codes to a minimum set of common ICU data elements. Each hospitalization was assigned a sequence of tokens, starting with a \texttt{BOS} beginning-of-sequence token. Tokens for age, sex, race, ethnicity, and admission type were inserted in this order at the time of admission. Interdepartmental transfers within the hospital received one token each for ingress and egress, at their respective times. Medications with dosages, labs with quantitative results, and vitals with measured values each received category-value tokenization. For each category (standardized kind medication, type of lab, type of vital), all numerical values reported within the training set corresponding to that category were used to determine decile cutoffs. All values for that category (in training, tuning, and held-out data) were then binned into \texttt{Q0}, \texttt{Q1}, \dots, \texttt{Q9} according to those deciles~\cite{Bur25}. For each category-value pair, a new ``fused'' token~\cite{Lee26,Guo26,Dwi24} was created (for example, \texttt{LAB-RES//so2\_arterial\_Q0} corresponds to a lab result for arterial SO$_2$ in the lowest decile). Tokens for medications and labs were inserted at the time of administration or measurement. For labs, a token corresponding to the lab was inserted at the time the lab was ordered, and a token corresponding to the lab category and its binned numerical value were inserted at the time the result was available. Continuous renal replacement therapy~\cite[CRRT, a.k.a. CKRT]{Tei25} with blood flow rate, patient assessments with numerical value where appropriate, and respiratory support~\cite{Hoh25} with FiO$_2$, peep, and tidal volume values also received category-value tokenization. Changes in code status~\cite{Ste17}, the practice of proning a patient~\cite{Gue20}, and discretized measurements of SOFA score~\cite{Sin16} were also tokenized. Additionally, the first occurrence of each event-based outcome (see Table~\ref{tbl:outcomes} for a complete list) was assigned a token and inserted at its respective time. All timelines end with an \texttt{EOS} end-of-sequence token. We did not use time-spacing or artificial time tokens~\cite{Pan21} as recent studies suggest they do not improve performance~\cite{Wor25,Att25}.

We first tokenized the MIMIC training set, learning both the 1344-token vocabulary and category-related decile cutoffs for binning. See Table~\ref{tbl:vocab} for a breakdown by token type and Table~\ref{tbl:tokens} for token-level statistics. We then froze the learned vocabulary and bins and applied the tokenizer to the tuning and held-out splits in MIMIC, and all splits in UCMC and NU.

\begin{table}[tb]
	\tbl{\textbf{Outcomes considered}: names, definitions, overall prevalence, and overall prevalence after 24 hours for each outcome. The following medication categories in CLIF-2.1 correspond to vasporessors: angiotensin, dopamine, epinephrine, norepinephrine, phenylephrine, and vasopressin.}
	{
		\begin{tabular}{llrr}
			\toprule
			name          & definition                                                & \begin{tabular}[c]{@{}r@{}}overall \\ prevalence\end{tabular} & \begin{tabular}[c]{@{}r@{}}post-24h \\ prev.\end{tabular} \\  \midrule
			anemia        & lab result: hemoglobin $<$7.0g/dL                         & 0.148                                                         & 0.093                                                     \\
			CRRT          & use of Continuous Renal Replacement Therapy               & 0.035                                                         & 0.024                                                     \\
			expired       & death prior to discharge                                  & 0.081                                                         & 0.081                                                     \\
			hyperkalemia  & lab result: potassium $\ge$6.5mEq/L                       & 0.040                                                         & 0.015                                                     \\
			hypernatremia & lab result: sodium $\ge$160mEq/L                          & 0.015                                                         & 0.009                                                     \\
			hypertension  & blood p.: systolic $\ge$180mmHg or diastolic $\ge$120mmHg & 0.397                                                         & 0.165                                                     \\
			hypokalemia   & lab result: potassium $<$2.5mEq/L                         & 0.014                                                         & 0.006                                                     \\
			hyponatremia  & lab result: sodium $<$120mEq/L                            & 0.016                                                         & 0.002                                                     \\
			hypotension   & mean arterial p. $<$65 mmHg or systolic $<$90mmHg         & 0.668                                                         & 0.174                                                     \\
			IMV           & use of Invasive Mechanical Ventilation                    & 0.367                                                         & 0.280                                                     \\
			tachycardia   & heart rate $\ge$130 beats per minute                      & 0.255                                                         & 0.119                                                     \\
			vasopressors      & administration of vasopressors                            & 0.391                                                         & 0.094                                                     \\
			\bottomrule
		\end{tabular}
	}
	\label{tbl:outcomes}
\end{table}

\subsection{Training}
All transformer models were trained from scratch on sequences of tokens from the EHR-specific vocabulary developed in the previous section. 
We initialized a version of the Llama-3.2 model architecture with a hidden size of 1024, intermediate size of 2048, 9 hidden layers, and 8 attention heads, for a total of 76.9M parameters.~\cite{Gra24}\footnote{Parameter size for tokenized event models tends to be significantly lower than for large language models. See Guo et al.'s recent review~\cite{Guo26b} for details.} We optimized with AdamW~\cite{Los19}, a variant of Adam~\cite{Kin15} with decoupled weight decay~\cite{Han88}. Training batches were formed by packing tokenized sequences into a $b\times 4096$-dimensional array in row-major order where $b=4$ is the batch size. Our standard training runs (GEM) trained on one epoch of data. We also completed extended training runs (GEM-$*$) on 5 epochs of data with time-based RoPE~\cite{Su24} and NEFTune~\cite{Jai24} ($\alpha=5$) for regularization.

\subsection{Federation}
 
We used FedAvg~\cite{McM17} to perform federated learning across datasets. For each dataset, a unique client trains a separate copy of the model on data from that dataset alone. At the end of each round, a central server collects the model weights from each client, forms a weighted average, and distributes the result back to the clients to serve as the initialization for the next round of training. We also tried FedAvgM~\cite{Hsu19} that adds momentum\footnote{Though Nesterov's accelerated gradient~\cite{Nes83} was originally proposed, we use Flower's~\cite{Beu20} implementation with Polyak's ``heavy ball'' momentum~\cite{Pol64}.} to FedAvg and FedAdam~\cite{Red21} that introduced adaptive updates to the weight aggregation process on the server. In our setup, all models are trained on one epoch of data. In the $n$-round setup, clients train on $1/n$th of their training dataset in each round.

\subsection{Evaluation}

\textbf{Inference.} We used representation-based inference. We truncated each sequence at the 24-hour mark and extracted the final hidden layer from a trained model as a vector-based representation associated to that sequence. On the training subset of each dataset, we then learned a logistic regression model to predict each outcome using the representations as features. Each regression used $L^2$ regularization at a strength chosen via 5-fold cross-validation to optimize AUC on the testing portions of the training set. Each outcome-specific model was then applied to the representations extracted from the held-out dataset to infer outcomes for each hospitalization in the held-out set.

\textbf{Metrics.} We consider 12 separate classification tasks, each corresponding to predicting if an outcome occurs after 24 hours, given that it did not occur within the first 24 hours. The 12 outcomes considered are those listed in Table~\ref{tbl:outcomes}. All patients must have been alive at the 24-hour threshold to have been included in the study. However, for all other outcomes, patients who had already experienced that outcome prior to the 24-hour threshold were excluded from that outcome-specific metric. We report both the area under the receiver-operator characteristic curve (ROC-AUC) and under the precision-recall curve (PR-AUC) for these tasks\footnote{For a thorough discussion of these metrics, see McDermott et al.~\cite{McD24}}. We compute confidence intervals for aggregated metrics using the standard bootstrapping approach~\cite[\S 13.3]{Efr93}, where each resampled dataset is used to calculate an average over each respective metric. To test the null hypothesis that two estimators have identical ROC-AUC performance against a two-sided alternative, we determine the empirical frequency at which the observed absolute difference in AUC exceeds that of the bootstrapped absolute difference in AUC of two sampled estimators~\cite[\emph{cf.} Algorithm 16.1]{Efr93}.

\textbf{Baselines.} We trained logistic regression (LR) classifiers with $L^2$ regularization on one-hot encoded indicators of each token from the first 24h for each patient. We also trained LightGBM~\cite{Ke17} (LGBM) models on token count-based 1-d vector representations of each patient. 

\section{Results}

\begin{table}[b]
    \renewcommand{\tablefont}{\scriptsize}
    \tbl{\textbf{Within-site, cross-site, and centralized multi-site performance:}
    ROC-AUC is shown on the left and PR-AUC on the right. Rows labeled UCMC,
    NU, or MIMIC indicate the dataset used for model training. Performance of
    models trained and evaluated at the same institution, using non-overlapping
    training and held-out subsets, is presented in boldface. Rows labeled
    \emph{All sites} report performance after centralized training on the pooled data (training split only)
    from all three institutions. 
    %Representation-based inference was used for the GEM models.
    }{
        \centering
        \begin{tabular}{llrrrrrr}
            \toprule
                &              &
                \multicolumn{3}{c}{held-out ROC-AUC} &
                \multicolumn{3}{c}{held-out PR-AUC} \\
            \cmidrule(lr){3-5}
            \cmidrule(lr){6-8}
            model
                & training set
                & UCMC
                & NU
                & MIMIC
                & UCMC
                & NU
                & MIMIC \\
            \midrule

            \multirow{4}{*}{LR}
                & UCMC
                & \textbf{0.746} (±0.013)
                & 0.690 (±0.013)
                & 0.675 (±0.012)
                & \textbf{0.323} (±0.013)
                & 0.196 (±0.006)
                & 0.195 (±0.008) \\

                & NU
                & 0.710 (±0.020)
                & \textbf{0.779} (±0.014)
                & 0.716 (±0.012)
                & 0.296 (±0.011)
                & \textbf{0.295} (±0.009)
                & 0.240 (±0.011) \\

                & MIMIC
                & 0.659 (±0.018)
                & 0.673 (±0.015)
                & \textbf{0.768} (±0.013)
                & 0.257 (±0.010)
                & 0.212 (±0.006)
                & \textbf{0.308} (±0.011) \\

                & \emph{All sites}
                & 0.756 (±0.018)
                & 0.790 (±0.012)
                & 0.781 (±0.013)
                & 0.343 (±0.012)
                & 0.307 (±0.011)
                & 0.319 (±0.013) \\

            \midrule

            \multirow{4}{*}{LGBM}
                & UCMC
                & \textbf{0.787} (±0.013)
                & 0.730 (±0.012)
                & 0.685 (±0.012)
                & \textbf{0.370} (±0.017)
                & 0.230 (±0.007)
                & 0.222 (±0.009) \\

                & NU
                & 0.723 (±0.021)
                & \textbf{0.793} (±0.014)
                & 0.707 (±0.014)
                & 0.310 (±0.012)
                & \textbf{0.320} (±0.012)
                & 0.239 (±0.009) \\

                & MIMIC
                & 0.707 (±0.018)
                & 0.735 (±0.012)
                & \textbf{0.800} (±0.012)
                & 0.282 (±0.010)
                & 0.253 (±0.007)
                & \textbf{0.345} (±0.011) \\

                & \emph{All sites}
                & 0.784 (±0.016)
                & 0.804 (±0.014)
                & 0.797 (±0.012)
                & 0.378 (±0.013)
                & 0.328 (±0.013)
                & 0.339 (±0.011) \\

            \midrule

            \multirow{4}{*}{GEM}
                & UCMC
                & \textbf{0.770} (±0.014)
                & 0.783 (±0.013)
                & 0.766 (±0.011)
                & \textbf{0.334} (±0.012)
                & 0.275 (±0.009)
                & 0.282 (±0.010) \\

                & NU
                & 0.770 (±0.013)
                & \textbf{0.794} (±0.014)
                & 0.772 (±0.011)
                & 0.335 (±0.013)
                & \textbf{0.297} (±0.011)
                & 0.294 (±0.011) \\

                & MIMIC
                & 0.750 (±0.015)
                & 0.769 (±0.015)
                & \textbf{0.798} (±0.011)
                & 0.318 (±0.012)
                & 0.271 (±0.009)
                & \textbf{0.312} (±0.012) \\

                & \emph{All sites}
                & 0.790 (±0.013)
                & 0.802 (±0.013)
                & 0.808 (±0.010)
                & 0.362 (±0.014)
                & 0.304 (±0.012)
                & 0.328 (±0.013) \\

            \midrule

            \multirow{4}{*}{GEM-$*$}
                & UCMC
                & \textbf{0.801} (±0.013)
                & 0.796 (±0.013)
                & 0.781 (±0.012)
                & \textbf{0.365} (±0.014)
                & 0.292 (±0.008)
                & 0.293 (±0.010) \\

                & NU
                & 0.776 (±0.015)
                & \textbf{0.809} (±0.013)
                & 0.790 (±0.011)
                & 0.357 (±0.014)
                & \textbf{0.315} (±0.012)
                & 0.311 (±0.011) \\

                & MIMIC
                & 0.784 (±0.012)
                & 0.784 (±0.013)
                & \textbf{0.821} (±0.010)
                & 0.348 (±0.012)
                & 0.287 (±0.011)
                & \textbf{0.344} (±0.012) \\

                & \emph{All sites}
                & 0.805 (±0.014)
                & 0.818 (±0.013)
                & 0.824 (±0.010)
                & 0.388 (±0.014)
                & 0.323 (±0.012)
                & 0.351 (±0.013) \\

            \bottomrule
        \end{tabular}
    }
    \label{tbl:xfer}
\end{table}

% We considered two conventional supervised baselines and two GEM training configurations. Logistic regression (LR) used binary indicators of observed tokens, while LightGBM (LGBM) used token counts. The standard GEM was pretrained for one epoch, matching the total local training budget used in the federated experiments. GEM-$*$ was pretrained for five epochs with NEFTune regularization and represents the optimized GEM configuration used for comparisons of predictive performance and transportability. For GEMs, we used representation-based inference: the pretrained model was held fixed, its final hidden states were used to represent each hospitalization, and outcome-specific logistic regression models were fit using these representations. We therefore use GEM-$*$ as the primary GEM comparator against LR and LGBM, while retaining the one-epoch GEM as the appropriate training-budget-matched reference for evaluating federated optimization.

The experiments address four related questions. First, we compare models trained and evaluated within the same institution with models transferred across institutions to assess the portability of the learned representations. Second, we train each model on the centrally pooled training data from all three institutions to quantify the potential benefit available from multi-site data when centralization is possible. Third, we compare this centralized one-epoch GEM with models trained using FedAvg, FedAvgM, and FedAdam, thereby isolating the performance difference associated with federated rather than centralized optimization. Finally, we vary the amount of local training data and the number of federated communication rounds to determine when multi-site training is most useful and how much communication is required. All models are evaluated separately on the original held-out population from each institution.

% We evaluated model performance across three independently held-out institutional datasets, comprising 12 post-24-hour clinical prediction tasks. For each task, models received the events observed during the first 24 hours of hospitalization and were evaluated on whether the outcome subsequently occurred. We report aggregate ROC-AUC and PR-AUC averaged across the 12 outcomes, with uncertainty estimated by bootstrap resampling.

\subsection{Generative pretraining produces more portable cross-site representations}

Within-site performance was strongest for the GEM-$*$ (Table~\ref{tbl:xfer}). GEM-$*$ achieved a mean ROC-AUC of 0.810 across the three institutional held-out sets, compared with 0.793 for LightGBM and 0.764 for LR. Its mean within-site PR-AUC was 0.341, similar to the 0.345 achieved by LightGBM. Thus, after extended training, GEM-derived representations improved ROC-AUC over the strong supervised baseline while providing comparable PR-AUC when complete local training data were available.

The advantage of GEM-$*$ was larger under cross-site transfer than for the non-GEM models. Mean cross-site ROC-AUC was 0.785 for GEM-$*$, compared with 0.715 for LightGBM and 0.687 for LR. Mean cross-site PR-AUC was 0.315 for GEM-$*$, compared with 0.256 and 0.233 for LightGBM and LR, respectively. GEM-$*$ therefore improved over LightGBM by 0.070 ROC-AUC and 0.059 PR-AUC when transferred to a different health system.

Table~\ref{tbl:xfer-cost} summarizes the corresponding ROC-AUC transfer penalties. GEM-$*$ incurred a mean penalty of 0.025 when transferred across institutions, compared with penalties of 0.079 for LightGBM and 0.077 for logistic regression. The same qualitative pattern was observed for PR-AUC, for which the transfer penalty was 0.027 for GEM-$*$ and 0.089 for LightGBM. The one-epoch GEM exhibited an even smaller transfer penalty, although its absolute within-site and cross-site performance was lower than that of GEM-$*$.

\begin{table}[b]
    \setlength{\tabcolsep}{12pt}
    \renewcommand{\tablefont}{\footnotesize}
    \tbl{\textbf{ROC-AUC transportability and benefit of centralized multi-site training.}
    Within-site performance is averaged over the three models trained and evaluated
    at the same institution, while cross-site performance is averaged over the six
    directed transfers. Transfer penalty is the difference between within-site and
    cross-site performance. Centralized gain is the difference between centralized
    multi-site and within-site performance.}{
        \centering
        \begin{tabular}{@{}lrrrr@{}}
            \toprule
            model
                & within-site
                & cross-site
                & transfer penalty
                & centralized gain \\
            \midrule
            LR
                & 0.764
                & 0.687
                & 0.077
                & 0.012 \\
            LGBM
                & 0.793
                & 0.715
                & 0.079
                & 0.002 \\
            GEM
                & 0.787
                & 0.768
                & \textbf{0.019}
                & \textbf{0.013} \\
            GEM-$*$
                & \textbf{0.810}
                & \textbf{0.785}
                & 0.025
                & 0.006 \\
            \bottomrule
        \end{tabular}
    }
    \label{tbl:xfer-cost}
\end{table}

\subsection{Centralized training yielded only modest gains}

Centralized multi-site training produced additional improvements, but the gains were modest relative to the increase in available training data (Tables \ref{tbl:xfer}, \ref{tbl:xfer-cost}). For GEM-$*$, centralized training increased mean ROC-AUC from 0.810 to 0.816 and mean PR-AUC from 0.341 to 0.354. The site-specific ROC-AUC improvements were 0.004 at UCMC, 0.009 at NU, and 0.003 at MIMIC. Centralized LightGBM increased mean ROC-AUC by only 0.002 relative to local training.

The optimized centralized GEM nevertheless achieved the highest ROC-AUC at each held-out institution, exceeding centralized LightGBM by 0.021 at UCMC, 0.014 at NU, and 0.027 at MIMIC. Its PR-AUC exceeded centralized LightGBM by 0.010 at UCMC and 0.012 at MIMIC, but was 0.005 lower at NU. Optimized GEMs therefore provided their clearest and most consistent advantage under cross-site transfer and in ROC-AUC. Neither generative pretraining nor centralized access to additional institutional data produced uniform gains across every metric and site.

These findings also separate data harmonization from predictive improvement. The shared CLIF representation enabled models to be trained, transferred, and evaluated consistently across independently governed health systems. Cross-site comparisons of quantile assignment (Figure~\ref{fig:cross-site-bins}) and GEM-derived estimates of cross-site cross-entropy (Figure~\ref{fig:cross-site-entropy}) are available in the appendix. Under the objectives and model configurations evaluated here, semantic alignment and increased data volume alone were not sufficient to produce large improvements over models trained using complete local data. Complete outcome-specific results are reported in Tables~\ref{tbl:roc-tkwz0}--\ref{tbl:pr-tkwz1}.

\subsection{Federated averaging approaches centralized multi-site performance}

We next compared federated GEM training with centralized models trained on pooled data from all three institutions (Table~\ref{tbl:fed}). To isolate the effect of federated optimization, all GEMs in this analysis used the same architecture and one-epoch training budget. Across the three institutional held-out sets, FedAvg was 0.011--0.024 lower in ROC-AUC and 0.012--0.035 lower in PR-AUC than GEM-all. FedAvgM produced slightly smaller deficits of 0.010--0.019 ROC-AUC and 0.008--0.027 PR-AUC. We did not detect a statistically significant difference between FedAvg and FedAvgM for any site or metric ($p>0.25$ for all six comparisons).

The federated models did not, however, consistently improve on models trained using all available local data. Relative to the corresponding one-epoch local GEM, FedAvgM differed by $+0.001$, $-0.002$, and $-0.008$ ROC-AUC at UCMC, NU, and MIMIC, respectively. The corresponding PR-AUC differences were $+0.003$, $-0.001$, and $-0.011$. FedAvg showed a similar pattern, matching the local GEM at UCMC but performing below it at NU and MIMIC.

FedAdam performed substantially worse than either averaging strategy, with deficits of 0.132--0.168 ROC-AUC and 0.065--0.119 PR-AUC relative to the centralized GEM. Simple parameter averaging, with or without server momentum, was therefore considerably more effective than the adaptive server-side optimizer in this setting.

The centralized supervised baselines provide context for the remaining federated gap. LR-all was below GEM-all in five of the six site-by-metric comparisons. LGBM-all was within 0.011 ROC-AUC of GEM-all at every site and exceeded it by 0.011--0.024 PR-AUC. The performance difference associated with replacing centralized GEM training with FedAvg or FedAvgM was therefore modest in absolute terms, but the one-epoch centralized GEM itself provided only a limited advantage over a strong supervised model trained on the same pooled data.

Overall, standard federated averaging reproduced much of the absolute performance of centralized GEM training, but did not substantively improve over complete data local training. These results suggest that, in this setting, the primary limitation was not the federated approach but primarily limitations in the efficacy of transfer learning between independent sites. This is important because it is often not possible to assess both centralized and federated learning in the same setting. 

\begin{table}[tb]
    \renewcommand{\tablefont}{\scriptsize}
    \tbl{\textbf{Performance of centralized baselines and federated GEMs relative to centralized GEM training.} The first row reports the absolute performance of a one-epoch GEM trained on centrally pooled data from all three institutions. All remaining rows report the difference relative to this GEM-all reference, with positive values indicating better performance. FedAvg, FedAvgM, and FedAdam were trained for 10 federated rounds. We did not detect a significant difference between FedAvg and FedAvgM for any site or metric ($p>0.25$ for all six comparisons).}{
        \centering
        \begin{tabular}{lrrrrrr}
            \toprule
                       & \multicolumn{3}{c}{held-out ROC-AUC}
                       & \multicolumn{3}{c}{held-out PR-AUC} \\
            \cmidrule(lr){2-4} \cmidrule(lr){5-7}
            strategy   & UCMC & NU & MIMIC & UCMC & NU & MIMIC \\
            \midrule

            GEM-all
                & 0.790 (±0.013)
                & 0.802 (±0.013)
                & 0.808 (±0.010)
                & 0.362 (±0.014)
                & 0.304 (±0.012)
                & 0.328 (±0.013) \\

            \midrule
            \multicolumn{7}{l}{\textit{Centralized supervised baselines:
            difference from GEM-all}} \\

            LR-all
                & -0.034
                & -0.012
                & -0.027
                & -0.019
                & \phantom{-}0.003
                & -0.009 \\

            LGBM-all
                & -0.006
                & \phantom{-}0.002
                & -0.011
                & \phantom{-}0.016
                & \phantom{-}0.024
                & \phantom{-}0.011 \\

            \midrule
            \multicolumn{7}{l}{\textit{Federated GEMs:
            difference from GEM-all}} \\

            FedAvg
                & -0.020
                & -0.011
                & -0.024
                & -0.025
                & -0.012
                & -0.035 \\

            FedAvgM
                & -0.019
                & -0.010
                & -0.018
                & -0.025
                & -0.008
                & -0.027 \\

            FedAdam
                & -0.160
                & -0.132
                & -0.168
                & -0.119
                & -0.065
                & -0.119 \\

            \bottomrule
        \end{tabular}
    }
    \label{tbl:fed}
\end{table}

\subsection{Multi-site models are most useful when local training data are limited}

We trained federated models using $n=1, 5, 10, 50$ rounds. For the $n=1$ case of a single federated round, this amounts to fully training a model for each site in parallel and then taking a weighted average over the weights. As $n$ grows, the amount of communication required between the server and the clients grows linearly. We find that performance benefits taper off at around 5-10 rounds of federated training. As cross-site communication is required after each round, this is a very encouraging practical finding.

\begin{figure}[tb]
	\centering
\centering
		\includegraphics[width=\textwidth]{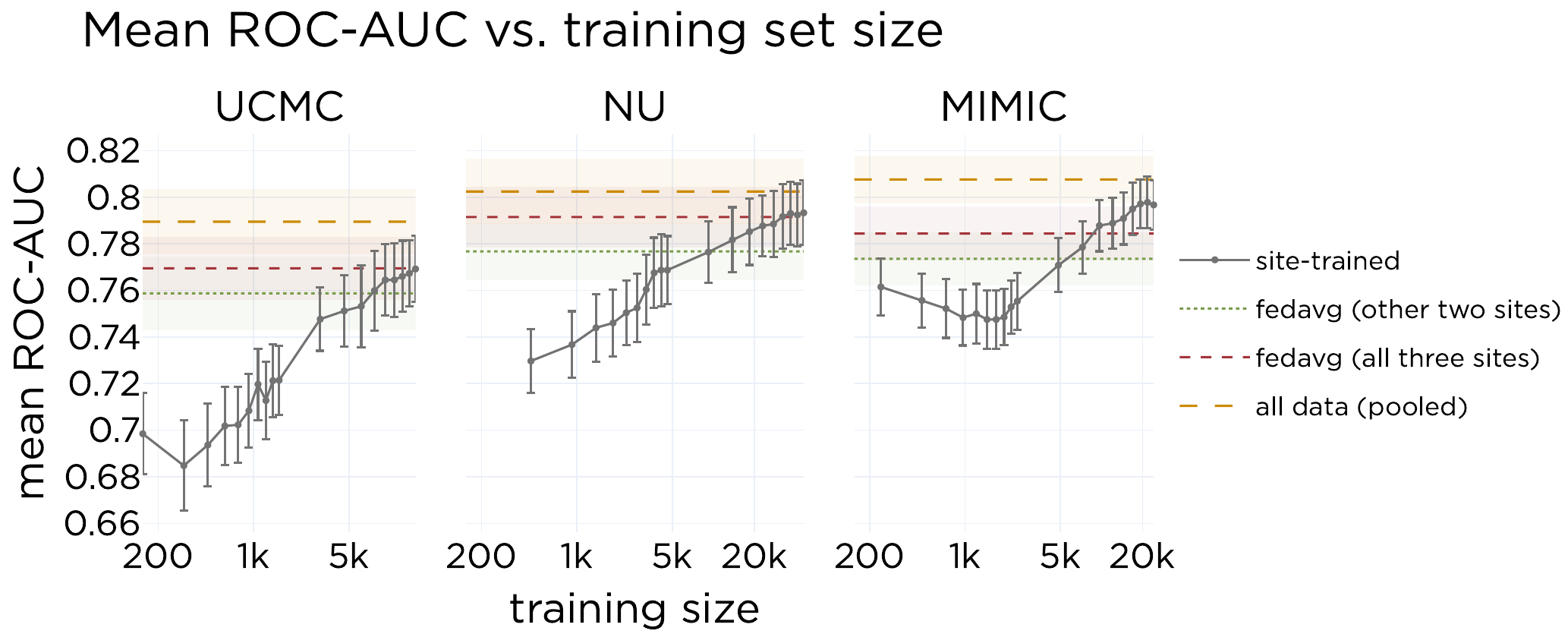}
		\caption{ROC-AUC performance on each respective fixed test set vs. number of training examples available. NU is the largest dataset and so has the widest graph. Error bars indicate 95\% confidence intervals for performance. Horizontal lines represent (in ascending order) performance of a federated model trained on the \emph{other} two sites, performance of a federated model trained on all sites, and performance of a model trained on a centralized aggregate dataset. See Figure~\ref{fig:local-pr-auc} for PR-AUC.}
		\label{fig:local-roc-auc}
\end{figure}

All prior comparisons used each institution's complete training set. We next varied the number of examples available for local GEM training while retaining the same fixed institutional held-out sets. Figures~\ref{fig:local-roc-auc} and~\ref{fig:local-pr-auc} compare the resulting local learning curves with the performance of a FedAvg model trained at the other two institutions, a FedAvg model trained across all three institutions, and a centralized GEM trained on pooled data.

At the smallest local sample sizes, the model trained federatively at the other two institutions generally outperformed the locally trained model. Local performance increased as additional institutional examples became available and eventually approached or exceeded the transferred two-site federated comparator. The crossover varied by institution and metric rather than occurring at a single universal sample size. Local models generally became competitive after several thousand examples, with some PR-AUC comparisons requiring more local data than the corresponding ROC-AUC comparisons.

The three-site federated and centralized models generally provided higher reference performance than the model trained only at the other two sites. Their advantage over local training nevertheless narrowed as the local sample size increased. At the largest local training sizes, the local learning curves approached the multi-site comparators, consistent with the limited benefit of federation observed when the complete institutional training sets were used.

These findings identify a conditional role for multi-site learning. A model learned from established institutions can provide useful initial performance for a newly participating or data-limited site. As sufficient local data accumulate, however, the incremental benefit of transferring a multi-site model decreases and local training becomes increasingly competitive.

\begin{table}[tb]
	\renewcommand{\tablefont}{\scriptsize}
	\tbl{\textbf{Aggregate performance vs. number of federated rounds for FedAvg}}{
		\centering
		\begin{tabular}{lrrrrrr}
			\toprule
			       & \multicolumn{3}{c}{held-out ROC-AUC} & \multicolumn{3}{c}{held-out PR-AUC}                                                                     \\
			\cmidrule(lr){2-4} \cmidrule(lr){5-7}
			rounds & UCMC                                 & NU                                  & MIMIC          & UCMC           & NU             & MIMIC          \\ \midrule
			1      & 0.759 (±0.014)                       & 0.779 (±0.014)                      & 0.764 (±0.012) & 0.311 (±0.011) & 0.281 (±0.011) & 0.281 (±0.010) \\
			5      & 0.767 (±0.014)                       & \textbf{0.792} (±0.013)          & 0.778 (±0.011) & 0.336 (±0.013) & \textbf{0.294} (±0.010) & 0.294 (±0.011) \\
			10     & \textbf{0.770} (±0.013)          & 0.790 (±0.013)                      & 0.784 (±0.012) & \textbf{0.338} (±0.013) & 0.291 (±0.012) & 0.294 (±0.010) \\
			50     & 0.767 (±0.016)                       & 0.788 (±0.014)                      & \textbf{0.787} (±0.010) & 0.332 (±0.012) & 0.292 (±0.011) & \textbf{0.296} (±0.011) \\
			\bottomrule
		\end{tabular}
	}
	\label{tbl:rounds}
\end{table}

\subsection{Most federated performance gains are obtained within 5--10 rounds}
Finally, we evaluated the relationship between the number of federated rounds and model performance. A one-round model trains independently at each institution for a complete epoch and averages the resulting weights only once. Increasing the number of rounds introduces more frequent communication and permits the shared model to be updated throughout local training.

Increasing the number of rounds from one to five improved mean ROC-AUC across the three sites from 0.767 to 0.779 and mean PR-AUC from 0.291 to 0.308 (Table~\ref{tbl:rounds}). Additional rounds provided little consistent benefit. Mean ROC-AUC was 0.781 after both 10 and 50 rounds, while mean PR-AUC was 0.308 after 10 rounds and 0.307 after 50 rounds. Although the site-specific optimum varied slightly, nearly all of the aggregate performance improvement was obtained within 5--10 rounds.
Because cross-site communication is required after every round, this saturation indicates that competitive federated GEM performance does not require frequent parameter exchange. We therefore used 10 rounds for comparisons among federated optimization strategies.

\section{Discussion}
This study separates three questions that are often conflated in multi-institutional EHR modeling: whether learned representations transfer across institutions, whether access to additional institutional data improves prediction, and whether federated optimization can recover the performance of centralized training. GEMs achieved the strongest cross-site performance and substantially smaller transfer penalties than logistic regression or LightGBM, while remaining competitive with the strongest baseline within individual sites. Federated averaging, in turn, approached the performance of centralized GEM training. However, centralized multi-site training itself provided only modest improvements over models trained using complete local datasets, and federated models did not consistently outperform their local counterparts. The clearest practical value of multi-site learning was therefore observed when local data were limited. Collectively, these results suggest that the principal challenge is not federated aggregation alone, but learning how to convert heterogeneous external data into reliable improvements at a particular target institution.

The limited gains from centralized training should not be interpreted as evidence that common data models or cross-institutional infrastructure are unnecessary. CLIF made it possible to train, transfer, and evaluate the same models across independently governed health systems using shared clinical semantics. This removes a major technical barrier and, importantly, allows the remaining statistical problem to be studied directly. Semantic harmonization does not eliminate differences in patient populations, clinical workflows, the frequency and/or directional biases in measurements,   treatment practices, outcome prevalence, or documentation density. Additional records may therefore contribute information that is weakly relevant, redundant, or even conflicting for a particular target site. A common data model should consequently be viewed as an enabling layer for multi-site learning, rather than a guarantee that pooled data will improve every model or institution.

These findings motivate approaches that explicitly optimize for target-site utility rather than a single globally averaged model. Promising directions include continued pretraining or lightweight adaptation at the target institution, personalized federated models that combine shared and site-specific parameters, and client weighting based on institutional similarity or measured transfer benefit. Mixture-of-experts models, site-conditioned representations, and selective transfer mechanisms could allow models to use external information only when it is relevant to the target population. More broadly, objectives designed to learn transportable clinical structure (e.g., patient health states), rather than only minimize aggregate next-token loss (i.e., care dynamics and documentation processes), may improve the value of heterogeneous data. Examples include invariance-aware representation learning, domain-adaptive pretraining, target-aware model selection, and training objectives that distinguish shared physiology from institution-specific patterns of care. Our results provide a benchmark to evaluate methods and suggest that improvements should be measured by target-site benefit.

Several limitations qualify these conclusions. We evaluated three health systems, all within adult critical care, and considered 12 event-based prediction tasks; results may differ across additional institutions, clinical domains, and outcomes. The vocabulary and numerical bins were learned from MIMIC and then fixed across sites, which enabled a common token space but may have affected institutions differently. We evaluated one transformer architecture and primarily used representation-based inference, so the findings may not generalize to other architectures, tokenization strategies, generative inference, or supervised fine-tuning. Predictive models in general require making a tremendous number of choices, and many of these are not yet fully characterized for GEMs and similar models. Decisions around dataset construction, to tokenization, model architecture and hyperparameter selection require additional exploration. The authors are incredibly grateful to the teams at Beth-Israel/MIT, Northwestern Medicine, and UCMC who have made data centralization possible for this experiment. However, federated learning typically involves a larger number of sites.  In the future, we plan to investigate the extent to which additional sites may improve federated performance.

\textbf{Data and code availability.} The MIMIC-IV-3.1 dataset~\cite{Joh23} is available in the CLIF standard to credentialed users on Physionet~\cite{Gol00} as ``MIMIC-IV-Ext-CLIF.'' The UCMC and NU datasets are available in the CLIF standard for federated, privacy-preserving analyses. Reasonable requests may be directed to WFP. Code to reproduce this work is available online: \url{https://github.com/bbj-lab/coreopsis}. Coreopsis relies on the \href{https://pypi.org/project/cocoa-tokenizer/}{\texttt{cocoa-tokenizer}} package for configurable collation and tokenization of tabular datasets and the \href{https://pypi.org/project/cotorra/}{\texttt{cotorra}} package for configurable training of generative event models, both of which are available in the \href{https://pypi.org}{Python package index}.

% This work was funded in part by the National Institutes of Health, specifically the National Institute of Neurological Disorders and Stroke grant R00NS114850 to BKB and National Library of Medicine grant R01LM014263 to WFP. 
\textbf{Acknowledgments.} Compute resources for this study were provided in part by the Center for Research Informatics (CRI) at the University of Chicago and funded by the Biological Sciences Division and the Institute for Translational Medicine/CTSA (NIH UL1TR002389).

\clearpage
%\bibliographystyle{ieeetr}
%\bibliography{coreopsis}

\begin{thebibliography}{10}

\bibitem{Kap20}
J.~Kaplan, S.~McCandlish, T.~Henighan, T.~B. Brown, B.~Chess, R.~Child,
  S.~Gray, A.~Radford, J.~Wu, and D.~Amodei, ``Scaling laws for neural language
  models.'' arXiv:2001.08361, 2020.

\bibitem{Hof22}
J.~Hoffmann, S.~Borgeaud, A.~Mensch, E.~Buchatskaya, T.~Cai, E.~Rutherford,
  D.~de~Las~Casas, L.~A. Hendricks, J.~Welbl, A.~Clark, T.~Hennigan, E.~Noland,
  K.~Millican, G.~van~den Driessche, B.~Damoc, A.~Guy, S.~Osindero,
  K.~Simonyan, E.~Elsen, O.~Vinyals, J.~Rae, and L.~Sifre, ``An empirical
  analysis of compute-optimal large language model training,'' in {\em Adv.
  Neur. Inf. Process. Syst.}, vol.~35, 2022.

\bibitem{Zha25}
S.~Zhang, Q.~Liu, N.~Usuyama, C.~Wong, T.~Naumann, and H.~Poon, ``Exploring
  scaling laws for {EHR} foundation models.'' arXiv:2505.22964, 2025.

\bibitem{Wax25}
S.~Waxler, P.~Blazek, D.~White, D.~Sneider, K.~Chung, M.~Nagarathnam,
  P.~Williams, H.~Voeller, K.~Wong, M.~Swanhorst, S.~Zhang, N.~Usuyama,
  C.~Wong, T.~Naumann, H.~Poon, A.~Loza, D.~Meeker, S.~Hain, and R.~Shah,
  ``Generative medical event models improve with scale.'' arXiv:2508.12104,
  2025.

\bibitem{Guo24}
L.~L. Guo, J.~Fries, E.~Steinberg, S.~L. Fleming, K.~Morse, C.~Aftandilian,
  J.~Posada, N.~Shah, and L.~Sung, ``A multi-center study on the adaptability
  of a shared foundation model for electronic health records,'' {\em npj Digit.
  Med.}, vol.~7, 2024.

\bibitem{Bur25}
M.~C. Burkhart, B.~Ramadan, Z.~Liao, K.~Chhikara, J.~C. Rojas, W.~F. Parker,
  and B.~K. Beaulieu-Jones, ``Foundation models for electronic health records:
  representation dynamics and transferability.'' arXiv:2504.10422, 2025.

\bibitem{Pan25}
C.~Pang, V.~Jeanselme, Y.~S. Choi, X.~Jiang, Z.~Jing, A.~Kashyap, Y.~Kobayashi,
  Y.~Li, F.~Pollet, K.~Natarajan, and S.~Joshi, ``{FoMoH}: A clinically
  meaningful foundation model evaluation for structured electronic health
  records.'' arXiv:2505.16941, 2025.

\bibitem{Mur10}
S.~N. Murphy, G.~Weber, M.~Mendis, V.~Gainer, H.~C. Chueh, S.~Churchill, and
  I.~Kohane, ``Serving the enterprise and beyond with informatics for
  integrating biology and the bedside (i2b2),'' {\em J. Am. Med. Inform.
  Assoc.}, vol.~17, no.~2, 2010.

\bibitem{Vos15}
E.~A. Voss, R.~Makadia, A.~Matcho, Q.~Ma, C.~Knoll, M.~Schuemie, F.~J. DeFalco,
  A.~Londhe, V.~Zhu, and P.~B. Ryan, ``Feasibility and utility of applications
  of the common data model to multiple, disparate observational health
  databases,'' {\em J. Am. Med. Inform. Assoc.}, vol.~22, no.~3, 2015.

\bibitem{Hon21}
C.~Hong, E.~Rush, M.~Liu, D.~Zhou, J.~Sun, A.~Sonabend, V.~M. Castro,
  P.~Schubert, V.~A. Panickan, T.~Cai, L.~Costa, Z.~He, N.~Link, R.~Hauser,
  J.~M. Gaziano, S.~N. Murphy, G.~Ostrouchov, Y.-L. Ho, E.~Begoli, J.~Lu,
  K.~Cho, K.~P. Liao, and T.~Cai, ``Clinical knowledge extraction via sparse
  embedding regression ({KESER}) with multi-center large scale electronic
  health record data,'' {\em npj Digit. Med.}, vol.~4, no.~1, 2021.

\bibitem{Zho26}
D.~Zhou, H.~Tong, L.~Wang, S.~Liu, X.~Xiong, Z.~Gan, G.~Romain, B.~P. Hejblum,
  Y.-C. Liu, C.~Hong, C.-L. Bonzel, T.~Cai, K.~Pan, Y.-L. Ho, L.~Costa,
  V.~A.~Panickan, J.~M. Gaziano, K.~D. Mandl, V.~Jouhet, R.~Thiebaut, Z.~Xia,
  K.~Cho, K.~Liao, and T.~Cai, ``Representation learning to advance
  multi-institutional studies with electronic health record data from {US} and
  {France},'' {\em Nat. Commun.}, vol.~17, 2026.

\bibitem{Nep22}
H.~T. Neprash, C.~C. McGlave, D.~A. Cross, B.~A. Virnig, M.~A. Puskarich, J.~D.
  Huling, A.~Z. Rozenshtein, and S.~S. Nikpay, ``Trends in ransomware attacks
  on {US} hospitals, clinics, and other health care delivery organizations,
  2016-2021,'' {\em JAMA Health Forum}, vol.~3, no.~12, 2022.

\bibitem{Jia25}
J.~X. Jiang, J.~S. Ross, and G.~Bai, ``Ransomware attacks and data breaches in
  {US} health care systems,'' {\em JAMA Netw. Open}, vol.~8, no.~5, 2025.

\bibitem{Roj25}
J.~C. Rojas, P.~G. Lyons, K.~Chhikara, V.~Chaudhari, S.~V. Bhavani, M.~Nour,
  K.~G. Buell, K.~D. Smith, C.~A. Gao, S.~Amagai, {\em et~al.}, ``A common
  longitudinal intensive care unit data format ({CLIF}) for critical illness
  research,'' {\em Intensive Care Med.}, vol.~51, 2025.

\bibitem{Lyo26}
P.~G. Lyons, K.~G. Buell, K.~A. Connell, M.~A. Christensen, C.~H. Hochberg,
  S.~Jain, W.~F. Parker, K.~Chhikara, J.~C. Rojas, C.~Blebea, S.~V. Bhavani,
  A.~K. Barker, N.~Mesfin, N.~E. Ingraham, and C.~A. Gao, ``Federation, not
  centralization: a new paradigm for electronic health record--based critical
  care research,'' {\em Ann. Am. Thorac. Soc.}, vol.~23, no.~4, 2026.

\bibitem{Bur26}
M.~C. Burkhart, B.~Ramadan, L.~Solo, W.~F. Parker, and B.~K. Beaulieu-Jones,
  ``Quantifying surprise in clinical care: Detecting highly informative events
  in electronic health records with foundation models,'' in {\em Pac. Symp.
  Biocomput.}, vol.~31, World Scientific, 2026.

\bibitem{Cha26}
P.~Chandak, G.~Kondas, L.~A. Friedman, I.~Kohane, and M.~McDermott,
  ``Everyquery: Zero-shot clinical prediction via task-conditioned pretraining
  over electronic health records.'' arXiv:2603.07900, 2026.

\bibitem{Guo26b}
L.~L. Guo, S.~E. Arciniegas, A.~P. Yan, J.~Fries, G.~A. Tomlinson, and L.~Sung,
  ``Systematic review of foundation models for structured electronic health
  records,'' {\em J. Am. Med. Inform. Assoc.}, vol.~33, no.~6, 2026.

\bibitem{Wor23}
M.~Wornow, R.~Thapa, E.~Steinberg, J.~A. Fries, and N.~Shah, ``{EHRSHOT}: An
  {EHR} benchmark for few-shot evaluation of foundation models,'' in {\em
  Neurips Datasets and Benchmarks Track}, vol.~36, 2023.

\bibitem{Wor25}
M.~Wornow, S.~Bedi, M.~A.~F. Hernandez, E.~Steinberg, J.~Fries, C.~R{\'e},
  O.~Koyejo, and N.~H. Shah, ``Context clues: Evaluating long context models
  for clinical prediction tasks on {EHRs},'' in {\em ICLR}, 2025.

\bibitem{Lee26}
I.~Lee, L.~Solo, M.~C. Burkhart, B.~Ramadan, W.~F. Parker, and B.~K.
  Beaulieu-Jones, ``Representation before training: A fixed-budget benchmark
  for generative medical event models.'' arXiv:2604.16775, 2026.

\bibitem{Guo26}
L.~L. Guo, S.~E. Arciniegas, J.~J. Lee, A.~P. Yan, G.~Tomlinson, J.~Fries, and
  L.~Sung, ``Tokenization tradeoffs in structured {EHR} foundation models.''
  arXiv:2603.15644, 2026.

\bibitem{Zha26}
A.~Zhang, T.~Ding, S.~J. Wagner, C.~Tian, M.~Y. Lu, R.~Pettit, J.~E. Lewis,
  A.~Misrahi, D.~Mo, L.~P. Le, and F.~Mahmood, ``A multimodal and temporal
  foundation model for virtual patient representations at healthcare system
  scale.'' arXiv:2604.18570, 2026.

\bibitem{McD23}
M.~McDermott, B.~Nestor, P.~Argaw, and I.~S. Kohane, ``Event stream {GPT}: A
  data pre-processing and modeling library for generative, pre-trained
  transformers over continuous-time sequences of complex events,'' in {\em Adv.
  Neur. Inf. Process. Syst.}, vol.~36, 2023.

\bibitem{Kra24}
Z.~Kraljevic, D.~Bean, A.~Shek, R.~Bendayan, H.~Hemingway, J.~A. Yeung,
  A.~Deng, A.~Baston, J.~Ross, E.~Idowu, J.~T. Teo, and R.~J.~B. Dobson,
  ``Foresight-a generative pretrained transformer for modelling of patient
  timelines using electronic health records: a retrospective modelling study,''
  {\em Lancet Digit. Health}, vol.~6, no.~4, 2024.

\bibitem{Ren24}
P.~Renc, Y.~Jia, A.~E. Samir, J.~Was, Q.~Li, D.~W. Bates, and A.~Sitek, ``Zero
  shot health trajectory prediction using transformer,'' {\em npj Digit. Med.},
  vol.~7, 2024.

\bibitem{Ren25}
P.~Renc, M.~K. Grzeszczyk, N.~Oufattole, D.~Goode, Y.~Jia, S.~Bieganski,
  M.~B.~A. McDermott, J.~Was, A.~E. Samir, J.~W. Cunningham, D.~W. Bates, and
  A.~Sitek, ``Foundation model of electronic medical records for adaptive risk
  estimation,'' {\em GigaScience}, vol.~14, 2025.

\bibitem{Sol26}
L.~Solo, M.~B.~A. McDermott, W.~F. Parker, B.~Ramadan, M.~C. Burkhart, and
  B.~K. Beaulieu-Jones, ``Efficient generative prediction for {EHR} foundation
  models: The {SCOPE} and {REACH} estimators.'' arXiv:2602.03730, 2026.

\bibitem{Ste24}
E.~Steinberg, J.~A. Fries, Y.~Xu, and N.~Shah, ``{MOTOR}: A time-to-event
  foundation model for structured medical records,'' in {\em ICLR}, 2024.

\bibitem{Fal25}
A.~Fallahpour, M.~Alinoori, W.~Ye, X.~Cao, A.~Afkanpour, and A.~Krishnan,
  ``{EHRMamba}: Towards generalizable and scalable foundation models for
  electronic health records,'' in {\em ML4H}, vol.~PMLR 259, 2025.

\bibitem{Teo24}
Z.~L. Teo, L.~Jin, N.~Liu, S.~Li, D.~Miao, X.~Zhang, W.~Y. Ng, T.~F. Tan, D.~M.
  Lee, K.~J. Chua, J.~Heng, Y.~Liu, R.~S. Mong~Goh, and D.~S. Wei~Ting,
  ``Federated machine learning in healthcare: A systematic review on clinical
  applications and technical architecture,'' {\em Cell Rep. Med.}, vol.~5,
  no.~3, 2024.

\bibitem{San26}
R.~Santos and P.~A. Keane, ``Federated learning authenticity standard for
  healthcare as derived from lessons in self-driving cars,'' {\em Commun.
  Med.}, 2026.

\bibitem{Joh23}
A.~E.~W. Johnson, L.~Bulgarelli, L.~Shen, A.~Gayles, A.~Shammout, S.~Horng,
  T.~J. Pollard, S.~Hao, B.~Moody, B.~Gow, L.-W.~H. Lehman, L.~A. Celi, and
  R.~G. Mark, ``{MIMIC}-{IV}, a freely accessible electronic health record
  dataset,'' {\em Sci. Data}, vol.~10, 2023.

\bibitem{Pol18}
T.~J. Pollard, A.~E.~W. Johnson, J.~D. Raffa, L.~A. Celi, R.~G. Mark, and
  O.~Badawi, ``The {eICU} collaborative research database, a freely available
  multi-center database for critical care research,'' {\em Sci. Data}, vol.~5,
  no.~1, 2018.

\bibitem{Dan22}
T.~K. Dang, X.~Lan, J.~Weng, and M.~Feng, ``Federated learning for electronic
  health records,'' {\em ACM Trans. Intell. Syst. Technol.}, vol.~13, no.~5,
  2022.

\bibitem{McM17}
B.~McMahan, E.~Moore, D.~Ramage, S.~Hampson, and B.~Ag\"uera~y Arcas,
  ``Communication-efficient learning of deep networks from decentralized
  data,'' in {\em AISTATS}, vol.~54, 2017.

\bibitem{Hsu19}
H.~Hsu, H.~Qi, and M.~Brown, ``Measuring the effects of non-identical data
  distribution for federated visual classification,'' in {\em Neurips Workshop
  on Federated Learning}, 2019.

\bibitem{Sho24}
O.~Ben~Shoham and N.~Rappoport, ``Federated learning of medical concepts
  embedding using {BEHRT},'' {\em JAMIA Open}, vol.~7, no.~4, 2024.

\bibitem{Ren25b}
P.~Renc, M.~K. Grzeszczyk, L.~Qian, N.~Oufattole, J.~Rasley, and A.~Sitek,
  ``Federated timeline synthesis: Scalable and private methodology for model
  training and deployment.'' arXiv:2506.23358, 2025.

\bibitem{Ove12}
J.~M. Overhage, P.~B. Ryan, C.~G. Reich, A.~G. Hartzema, and P.~E. Stang,
  ``Validation of a common data model for active safety surveillance
  research,'' {\em J. Am. Med. Inform. Assoc.}, vol.~19, no.~1, 2012.

\bibitem{Kim26}
J.~Kim, J.~Kim, K.~Hur, and E.~Choi, ``Federated learning for heterogeneous
  electronic health record systems with cost effective participant selection.''
  arXiv:2404.13318, 2026.

\bibitem{Guo26c}
L.~L. Guo, A.~P. Yan, E.~Vettese, and L.~Sung, ``{PORTER}: Language-grounded
  event representations for portable structured ehr foundation models.''
  arXiv:2606.24102, 2026.

\bibitem{Dwi24}
V.~P. Dwivedi, V.~Schlegel, A.~T. Liu, T.-T. Nguyen, A.~R. Kashyap, J.~Wei,
  W.-H. Yin, S.~Winkler, and R.~T. Tan, ``Representation learning of structured
  data for medical foundation models,'' in {\em UniReps}, vol.~PMLR 285, 2024.

\bibitem{Tei25}
J.~P. Teixeira, S.~Hiremath, A.~O. Kabli, O.~G. Rewa, and E.~G. Clark,
  ``Continuous kidney replacement therapies: Core curriculum 2025,'' {\em Am.
  J. Kidney Dis.}, vol.~85, no.~6, 2025.

\bibitem{Hoh25}
F.~Hohmann, F.~Fichtner, T.~Becher, D.~Schaedler, C.~Putensen, T.~Muders,
  I.~Schroeder, C.~Karagiannidis, H.~Wrigge, D.~Berger, M.~Grupp, F.~Grundeis,
  V.~Buenger, A.~Sachkova, S.~Henkel, M.~Habicher, M.~Sander, S.~Laudi,
  S.~Weber-Carstens, and O.~Moerer, ``Clinical guideline for treating acute
  respiratory insufficiency with invasive ventilation and extracorporeal
  membrane oxygenation: Updated evidence-based recommendations for choosing
  modes and setting parameters of mechanical ventilation,'' {\em Respiration},
  vol.~105, no.~7, 2025.

\bibitem{Ste17}
E.~K. Stevenson, H.~M. Mehter, A.~J. Walkey, and R.~S. Wiener, ``Association
  between do not resuscitate/do not intubate status and resident physician
  decision-making: A national survey,'' {\em Ann. Am. Thorac. Soc.}, vol.~14,
  no.~4, 2017.

\bibitem{Gue20}
C.~Gu{\'e}rin, R.~K. Albert, J.~Beitler, L.~Gattinoni, S.~Jaber, J.~J. Marini,
  L.~Munshi, L.~Papazian, A.~Pesenti, A.~Vieillard-Baron, and J.~Mancebo,
  ``Prone position in {ARDS} patients: why, when, how and for whom,'' {\em
  Intensive Care Med.}, vol.~46, no.~12, 2020.

\bibitem{Sin16}
M.~Singer, C.~S. Deutschman, C.~W. Seymour, M.~Shankar-Hari, D.~Annane,
  M.~Bauer, R.~Bellomo, G.~R. Bernard, J.-D. Chiche, C.~M. Coopersmith, R.~S.
  Hotchkiss, M.~M. Levy, J.~C. Marshall, G.~S. Martin, S.~M. Opal, G.~D.
  Rubenfeld, T.~van~der Poll, J.-L. Vincent, and D.~C. Angus, ``The third
  international consensus definitions for sepsis and septic shock
  ({Sepsis-3}),'' {\em JAMA}, vol.~315, 02 2016.

\bibitem{Pan21}
C.~Pang, X.~Jiang, K.~S. Kalluri, M.~Spotnitz, R.~Chen, A.~Perotte, and
  K.~Natarajan, ``{CEHR-BERT}: Incorporating temporal information from
  structured {EHR} data to improve prediction tasks,'' in {\em ML4H}, 2021.

\bibitem{Att25}
R.~A. Attrach, R.~Fani, D.~Restrepo, Y.~Jia, and P.~Sch{\"u}ffler, ``Rethinking
  tokenization for clinical time series: When less is more,'' in {\em ML4H},
  2025.

\bibitem{Gra24}
A.~Grattafiori, A.~Dubey, A.~Jauhri, A.~Pandey, A.~Kadian, A.~Al-Dahle, {\em
  et~al.}, ``The {Llama} 3 herd of models.'' arXiv 2407.21783, 2024.

\bibitem{Los19}
I.~Loshchilov and F.~Hutter, ``Decoupled weight decay regularization,'' in {\em
  ICLR}, 2019.

\bibitem{Kin15}
D.~P. Kingma and J.~Ba, ``Adam: A method for stochastic optimization,'' in {\em
  ICLR}, 2015.

\bibitem{Han88}
S.~Hanson and L.~Pratt, ``Comparing biases for minimal network construction
  with back-propagation,'' in {\em Adv. Neur. Inf. Process. Syst.}, 1988.

\bibitem{Su24}
J.~Su, M.~Ahmed, Y.~Lu, S.~Pan, W.~Bo, and Y.~Liu, ``{RoFormer}: Enhanced
  transformer with rotary position embedding,'' {\em Neurocomput.}, vol.~568,
  2024.

\bibitem{Jai24}
N.~Jain, P.~yeh Chiang, Y.~Wen, J.~Kirchenbauer, H.-M. Chu, G.~Somepalli, B.~R.
  Bartoldson, B.~Kailkhura, A.~Schwarzschild, A.~Saha, M.~Goldblum, J.~Geiping,
  and T.~Goldstein, ``{NEFT}une: Noisy embeddings improve instruction
  finetuning,'' in {\em ICLR}, 2024.

\bibitem{Nes83}
Y.~E. Nesterov, ``A method of solving a convex programming problem with
  convergence rate o(1/sqr(k)),'' {\em Sov. Math. Dokl.}, vol.~27, 1983.

\bibitem{Beu20}
D.~J. Beutel, T.~Topal, A.~Mathur, X.~Qiu, J.~Fernandez-Marques, Y.~Gao,
  L.~Sani, K.~H. Li, T.~Parcollet, P.~P.~B. de~Gusm{\~a}o, and N.~D. Lane,
  ``Flower: A friendly federated learning research framework.''
  arXiv:2007.14390, 2020.

\bibitem{Pol64}
B.~T. Polyak, ``Some methods of speeding up the convergence of iteration
  methods,'' {\em USSR Comput. Math. Math. Phys.}, vol.~4, no.~5, 1964.

\bibitem{Red21}
S.~Reddi, Z.~Charles, M.~Zaheer, Z.~Garrett, K.~Rush, J.~Kone{\v c}n{\'y},
  S.~Kumar, and H.~B. McMahan, ``Adaptive federated optimization,'' in {\em
  ICLR}, 2021.

\bibitem{McD24}
M.~B. McDermott, H.~Zhang, L.~H. Hansen, G.~Angelotti, and J.~Gallifant, ``A
  closer look at {AUROC} and {AUPRC} under class imbalance,'' in {\em Adv.
  Neur. Inf. Proc. Sys.}, 2024.

\bibitem{Efr93}
B.~Efron and R.~J. Tibshirani, {\em An Introduction to the Bootstrap}, vol.~57
  of {\em Monographs on Statistics and Applied Probability}.
\newblock Chapman and Hall, 1993.

\bibitem{Ke17}
G.~Ke, Q.~Meng, T.~Finley, T.~Wang, W.~Chen, W.~Ma, Q.~Ye, and T.-Y. Liu,
  ``Lightgbm: A highly efficient gradient boosting decision tree,'' in {\em
  Adv. Neur. Inf. Proc. Sys.}, vol.~30, Curran Associates, Inc., 2017.

\bibitem{Gol00}
A.~L. Goldberger, L.~A.~N. Amaral, L.~Glass, J.~M. Hausdorff, P.~C. Ivanov,
  R.~G. Mark, J.~E. Mietus, G.~B. Moody, C.-K. Peng, and H.~E. Stanley,
  ``Physiobank, physiotoolkit, and physionet,'' {\em Circulation}, vol.~101,
  no.~23, 2000.

\end{thebibliography}

\clearpage
\renewcommand{\theHtable}{app.\thetable}%
\renewcommand{\theHfigure}{app.\thefigure}%
\appendix{Supplementary tables}

\begin{table}[tbh]
	\tbl{\textbf{Breakdown of the 1344-token vocabulary}}
	{
		\begin{tabular}{lllr}
			\toprule
			category & description         & example token(s)                                                    & count \\
			\midrule
			ADMN     & admission type      & \texttt{ADMN//direct}, \texttt{ADMN//elective}                      & 3     \\
			AGE      & age                 & \texttt{AGE//age\_Q0}, \texttt{AGE//age\_Q5}, \texttt{AGE//age\_Q9} & 10    \\
			ASMT     & assessment          & \texttt{ASMT//gcs\_eye\_Q1}, \texttt{ASMT//rass\_Q4}                & 76    \\
			CODE     & code status         & \texttt{CODE//dnr}, \texttt{CODE//full}                             & 4     \\
			CRRT     & dialysis            & \texttt{CRRT//cvvh\_Q0}, \texttt{CRRT//cvvhd\_Q4}                   & 18    \\
			DSCG     & discharge           & \texttt{DSCG//acute\_care\_hospital}, \texttt{DSCG//home}           & 12    \\
			ETHN     & ethnicity           & \texttt{ETHN//hispanic}, \texttt{ETHN//unknown}                     & 3     \\
			LAB-ORD  & lab order           & \texttt{LAB-ORD//albumin}, \texttt{LAB-ORD//wbc}                    & 45    \\
			LAB-RES  & lab result          & \texttt{LAB-RES//albumin\_Q0}, \texttt{LAB-RES//alt\_Q2}            & 474   \\
			LABEL    & additional labels   & \texttt{LABEL//anemia\_init}, \texttt{LABEL//crrt\_init}            & 11    \\
			MED-CTS  & continuous med.     & \texttt{MED-CTS//dopamine\_Q8}                                      & 363   \\
			MED-INT  & intermittent med.   & \texttt{MED-INT//adenosine\_Q7}                                     & 153   \\
			POSN     & proning             & \texttt{POSN//prone}                                                & 1     \\
			RACE     & race                & \texttt{RACE//asian}, \texttt{RACE//white}                          & 7     \\
			RESP     & respiratory support & \texttt{RESP//fio2\_set\_Q1}                                        & 34    \\
			SEX      & sex                 & \texttt{SEX//female}, \texttt{SEX//male}                            & 2     \\
			SOFA     & sofa score          & \texttt{SOFA//cns-0}, \texttt{SOFA//coag-0}                         & 23    \\
			VTL      & vitals              & \texttt{VTL//dbp\_Q0}, \texttt{VTL//weight\_kg\_Q9}                 & 88    \\
			XFR-IN   & transfer in         & \texttt{XFR-IN//ed}, \texttt{XFR-IN//psych}                         & 7     \\
			XFR-OUT  & transfer out        & \texttt{XFR-OUT//ed}, \texttt{XFR-OUT//psych}                       & 7     \\
			special  & miscellaneous       & \texttt{BOS}, \texttt{EOS}, \texttt{UNK}                            & 3     \\
			\bottomrule
		\end{tabular}
	}
	\label{tbl:vocab}
\end{table}

\begin{table}[tbh]
	\tbl{\textbf{Token-based statistics}: We report the total number of tokens available for training in each dataset, followed by the number of tokens per patient within the training set. We then report the average number of tokens occurring prior to the 24-hour threshold in the held-out (test) datasets. This will correspond to the number of tokens the model processes prior to making a prediction.}
	{
		\begin{tabular}{lrrr}
			\toprule
			        & \multicolumn{2}{c}{training set} & \multicolumn{1}{c}{held-out set}                            \\
			\cmidrule(lr){2-3} \cmidrule(lr){4-4}
			site    & total tokens                     & tokens/patient & avg. tokens $<24$h \\ \midrule
			UCMC    & 59.9M                            & 3.9k           & 651.9              \\
			NU      & 108.0M                           & 2.3k           & 537.9              \\
			MIMIC   & 46.4M                            & 1.9k           & 355.1              \\ \midrule
			overall & 214.3M                           & 2.5k           & 506.8              \\
			\bottomrule
		\end{tabular}
	}
	\label{tbl:tokens}
\end{table}

\begin{table}[tbh]
	\renewcommand{\tablefont}{\scriptsize}
	\tbl{\textbf{ROC-AUC performance by outcome (first six)}: we report ROC-AUC for each of the first six outcomes (anemia, CRRT, expired, hyperkalemia, hypernatremia, and hypertension)}{
		\centering
		\begin{tabular}{@{}c@{}}
			\begin{tabular}{llrrrrrrrrr}
				\toprule
				                              &         & \multicolumn{3}{c}{anemia} & \multicolumn{3}{c}{CRRT} & \multicolumn{3}{c}{expired}                                                 \\
				\cmidrule(lr){3-5} \cmidrule(lr){6-8} \cmidrule(lr){9-11}
				                              & outcome & UCMC                       & NU                       & MIMIC                       & UCMC  & NU    & MIMIC & UCMC  & NU    & MIMIC \\
				\midrule
				\multirow[c]{4}{*}{UCMC}      & LR      & 0.809                      & 0.749                    & 0.708                       & 0.805 & 0.737 & 0.823 & 0.904 & 0.821 & 0.793 \\
				                              & LGBM    & 0.848                      & 0.805                    & 0.779                       & 0.876 & 0.841 & 0.851 & 0.918 & 0.858 & 0.810 \\
				                              & GEM     & 0.796                      & 0.773                    & 0.749                       & 0.833 & 0.844 & 0.857 & 0.890 & 0.880 & 0.847 \\
				                              & GEM-$*$ & 0.835                      & 0.819                    & 0.790                       & 0.871 & 0.872 & 0.886 & 0.906 & 0.892 & 0.855 \\
				\midrule
				\multirow[c]{4}{*}{NU}        & LR      & 0.746                      & 0.819                    & 0.774                       & 0.812 & 0.866 & 0.848 & 0.863 & 0.937 & 0.818 \\
				                              & LGBM    & 0.780                      & 0.841                    & 0.782                       & 0.859 & 0.886 & 0.878 & 0.873 & 0.949 & 0.808 \\
				                              & GEM     & 0.793                      & 0.808                    & 0.757                       & 0.842 & 0.858 & 0.861 & 0.886 & 0.892 & 0.855 \\
				                              & GEM-$*$ & 0.827                      & 0.840                    & 0.807                       & 0.864 & 0.876 & 0.891 & 0.899 & 0.907 & 0.872 \\
				\midrule
				\multirow[c]{4}{*}{MIMIC}     & LR      & 0.691                      & 0.742                    & 0.811                       & 0.818 & 0.760 & 0.873 & 0.863 & 0.819 & 0.889 \\
				                              & LGBM    & 0.752                      & 0.790                    & 0.840                       & 0.841 & 0.824 & 0.909 & 0.887 & 0.899 & 0.898 \\
				                              & GEM     & 0.786                      & 0.772                    & 0.787                       & 0.836 & 0.853 & 0.893 & 0.882 & 0.879 & 0.867 \\
				                              & GEM-$*$ & 0.802                      & 0.791                    & 0.837                       & 0.861 & 0.860 & 0.915 & 0.905 & 0.884 & 0.895 \\
				\midrule
				\multirow[c]{3}{*}{federated} & FedAvg  & 0.796                      & 0.791                    & 0.771                       & 0.839 & 0.857 & 0.877 & 0.888 & 0.887 & 0.861 \\
				                              & FedAvgM & 0.801                      & 0.798                    & 0.778                       & 0.843 & 0.860 & 0.880 & 0.886 & 0.888 & 0.863 \\
				                              & FedAdam & 0.663                      & 0.652                    & 0.622                       & 0.629 & 0.687 & 0.667 & 0.733 & 0.735 & 0.640 \\
				\midrule
				\multirow[c]{4}{*}{all}       & LR      & 0.826                      & 0.824                    & 0.823                       & 0.873 & 0.880 & 0.897 & 0.914 & 0.936 & 0.892 \\
				                              & LGBM    & 0.841                      & 0.841                    & 0.839                       & 0.877 & 0.892 & 0.902 & 0.918 & 0.948 & 0.901 \\
				                              & GEM     & 0.824                      & 0.827                    & 0.815                       & 0.856 & 0.872 & 0.910 & 0.904 & 0.896 & 0.879 \\
				                              & GEM-$*$ & 0.846                      & 0.843                    & 0.841                       & 0.886 & 0.890 & 0.921 & 0.920 & 0.911 & 0.893 \\
				\bottomrule                                                                                                                                                                   \\
			\end{tabular}                     \\ \\

			\begin{tabular}{llrrrrrrrrr}
				\toprule
				                              &         & \multicolumn{3}{c}{hyperkalemia} & \multicolumn{3}{c}{hypernatremia} & \multicolumn{3}{c}{hypertension}                                                 \\
				\cmidrule(lr){3-5} \cmidrule(lr){6-8} \cmidrule(lr){9-11}
				                              & outcome & UCMC                             & NU                                & MIMIC                            & UCMC  & NU    & MIMIC & UCMC  & NU    & MIMIC \\
				\midrule
				\multirow[c]{4}{*}{UCMC}      & LR      & 0.648                            & 0.677                             & 0.622                            & 0.708 & 0.714 & 0.711 & 0.699 & 0.629 & 0.625 \\
				                              & LGBM    & 0.677                            & 0.720                             & 0.569                            & 0.820 & 0.768 & 0.684 & 0.733 & 0.664 & 0.675 \\
				                              & GEM     & 0.731                            & 0.768                             & 0.706                            & 0.827 & 0.845 & 0.811 & 0.724 & 0.734 & 0.767 \\
				                              & GEM-$*$ & 0.764                            & 0.794                             & 0.725                            & 0.845 & 0.849 & 0.826 & 0.745 & 0.747 & 0.777 \\
				\midrule
				\multirow[c]{4}{*}{NU}        & LR      & 0.586                            & 0.707                             & 0.600                            & 0.690 & 0.828 & 0.687 & 0.667 & 0.738 & 0.695 \\
				                              & LGBM    & 0.628                            & 0.769                             & 0.626                            & 0.761 & 0.870 & 0.684 & 0.696 & 0.753 & 0.693 \\
				                              & GEM     & 0.720                            & 0.772                             & 0.721                            & 0.810 & 0.864 & 0.843 & 0.721 & 0.749 & 0.777 \\
				                              & GEM-$*$ & 0.748                            & 0.799                             & 0.735                            & 0.806 & 0.885 & 0.825 & 0.732 & 0.759 & 0.780 \\
				\midrule
				\multirow[c]{4}{*}{MIMIC}     & LR      & 0.565                            & 0.577                             & 0.685                            & 0.637 & 0.658 & 0.725 & 0.596 & 0.663 & 0.773 \\
				                              & LGBM    & 0.665                            & 0.686                             & 0.652                            & 0.725 & 0.721 & 0.819 & 0.636 & 0.691 & 0.793 \\
				                              & GEM     & 0.703                            & 0.757                             & 0.730                            & 0.800 & 0.819 & 0.863 & 0.714 & 0.733 & 0.788 \\
				                              & GEM-$*$ & 0.733                            & 0.771                             & 0.732                            & 0.823 & 0.838 & 0.868 & 0.722 & 0.741 & 0.807 \\
				\midrule
				\multirow[c]{3}{*}{federated} & FedAvg  & 0.733                            & 0.783                             & 0.718                            & 0.824 & 0.858 & 0.829 & 0.725 & 0.741 & 0.782 \\
				                              & FedAvgM & 0.725                            & 0.779                             & 0.725                            & 0.822 & 0.854 & 0.827 & 0.725 & 0.744 & 0.781 \\
				                              & FedAdam & 0.633                            & 0.613                             & 0.583                            & 0.697 & 0.657 & 0.621 & 0.606 & 0.655 & 0.669 \\
				\midrule
				\multirow[c]{4}{*}{all}       & LR      & 0.657                            & 0.774                             & 0.707                            & 0.768 & 0.867 & 0.775 & 0.716 & 0.742 & 0.774 \\
				                              & LGBM    & 0.685                            & 0.774                             & 0.670                            & 0.816 & 0.887 & 0.847 & 0.733 & 0.757 & 0.795 \\
				                              & GEM     & 0.745                            & 0.785                             & 0.733                            & 0.838 & 0.877 & 0.872 & 0.739 & 0.752 & 0.791 \\
				                              & GEM-$*$ & 0.758                            & 0.812                             & 0.742                            & 0.847 & 0.895 & 0.880 & 0.756 & 0.764 & 0.803 \\
				\bottomrule                                                                                                                                                                                       \\
			\end{tabular} \\
		\end{tabular}
	}
	\label{tbl:roc-tkwz0}
\end{table}

%%%%%%

\begin{table}[tbh]
	\renewcommand{\tablefont}{\scriptsize}
	\tbl{\textbf{ROC-AUC performance by outcome (last six)}: we report ROC-AUC for each of the last six outcomes (hypokalemia, hyponatremia, hypotension, IMV, tachycardia, and vasopressors).}{
		\centering
		\begin{tabular}{@{}c@{}}

			\begin{tabular}{llrrrrrrrrr}
				\toprule
				                              &               & \multicolumn{3}{c}{hypokalemia} & \multicolumn{3}{c}{hyponatremia} & \multicolumn{3}{c}{hypotension}                                                 \\
				\cmidrule(lr){3-5} \cmidrule(lr){6-8} \cmidrule(lr){9-11}
				                              & outcome       & UCMC                            & NU                               & MIMIC                           & UCMC  & NU    & MIMIC & UCMC  & NU    & MIMIC \\
				\midrule
				\multirow[c]{4}{*}{UCMC}      & LR            & 0.598                           & 0.477                            & 0.617                           & 0.829 & 0.694 & 0.536 & 0.727 & 0.685 & 0.694 \\
				                              & LGBM          & 0.621                           & 0.541                            & 0.552                           & 0.813 & 0.659 & 0.562 & 0.770 & 0.697 & 0.664 \\
				                              & GEM           & 0.631                           & 0.704                            & 0.682                           & 0.762 & 0.719 & 0.596 & 0.762 & 0.747 & 0.817 \\
				                              & GEM-$*$       & 0.668                           & 0.704                            & 0.687                           & 0.868 & 0.703 & 0.625 & 0.773 & 0.764 & 0.825 \\
				\midrule
				\multirow[c]{4}{*}{NU}        & LR            & 0.622                           & 0.670                            & 0.600                           & 0.624 & 0.646 & 0.686 & 0.715 & 0.747 & 0.720 \\
				                              & LGBM          & 0.583                           & 0.649                            & 0.568                           & 0.515 & 0.592 & 0.566 & 0.721 & 0.768 & 0.740 \\
				                              & GEM           & 0.609                           & 0.704                            & 0.685                           & 0.859 & 0.703 & 0.617 & 0.742 & 0.766 & 0.819 \\
				                              & GEM-$*$       & 0.616                           & 0.721                            & 0.673                           & 0.739 & 0.695 & 0.666 & 0.757 & 0.781 & 0.838 \\
				\midrule
				\multirow[c]{4}{*}{MIMIC}     & LR            & 0.592                           & 0.500                            & 0.701                           & 0.526 & 0.562 & 0.490 & 0.642 & 0.662 & 0.816 \\
				                              & LGBM          & 0.563                           & 0.590                            & 0.690                           & 0.603 & 0.642 & 0.649 & 0.686 & 0.713 & 0.845 \\
				                              & GEM           & 0.605                           & 0.684                            & 0.702                           & 0.721 & 0.627 & 0.697 & 0.736 & 0.742 & 0.836 \\
				                              & GEM-$*$       & 0.660                           & 0.659                            & 0.703                           & 0.838 & 0.715 & 0.754 & 0.758 & 0.756 & 0.856 \\
				\midrule
				\multirow[c]{3}{*}{federated} & FedAvg  0.615 & 0.703                           & 0.690                            & 0.801                           & 0.708 & 0.671 & 0.747 & 0.762 & 0.822         \\
				                              & FedAvgM       & 0.631                           & 0.705                            & 0.707                           & 0.788 & 0.705 & 0.698 & 0.754 & 0.761 & 0.824 \\
				                              & FedAdam       & 0.557                           & 0.613                            & 0.567                           & 0.573 & 0.580 & 0.563 & 0.604 & 0.678 & 0.744 \\
				\midrule
				\multirow[c]{4}{*}{all}       & LR            & 0.632                           & 0.655                            & 0.725                           & 0.649 & 0.630 & 0.558 & 0.736 & 0.754 & 0.801 \\
				                              & LGBM          & 0.649                           & 0.659                            & 0.689                           & 0.737 & 0.658 & 0.619 & 0.770 & 0.771 & 0.836 \\
				                              & GEM           & 0.624                           & 0.707                            & 0.715                           & 0.848 & 0.703 & 0.712 & 0.768 & 0.774 & 0.840 \\
				                              & GEM-$*$       & 0.643                           & 0.738                            & 0.711                           & 0.835 & 0.711 & 0.754 & 0.783 & 0.783 & 0.860 \\
				\bottomrule                                                                                                                                                                                          \\
			\end{tabular} \\ \\

			\begin{tabular}{llrrrrrrrrr}
				\toprule
				                              &         & \multicolumn{3}{c}{IMV} & \multicolumn{3}{c}{tachycardia} & \multicolumn{3}{c}{vasopressors}                                                 \\
				\cmidrule(lr){3-5} \cmidrule(lr){6-8} \cmidrule(lr){9-11}
				                              & outcome & UCMC                    & NU                              & MIMIC                            & UCMC  & NU    & MIMIC & UCMC  & NU    & MIMIC \\
				\midrule
				\multirow[c]{4}{*}{UCMC}      & LR      & 0.783                   & 0.730                           & 0.662                            & 0.696 & 0.662 & 0.660 & 0.755 & 0.709 & 0.649 \\
				                              & LGBM    & 0.839                   & 0.768                           & 0.714                            & 0.734 & 0.670 & 0.704 & 0.793 & 0.773 & 0.668 \\
				                              & GEM     & 0.807                   & 0.836                           & 0.799                            & 0.735 & 0.722 & 0.771 & 0.748 & 0.830 & 0.791 \\
				                              & GEM-$*$ & 0.822                   & 0.845                           & 0.800                            & 0.750 & 0.734 & 0.774 & 0.777 & 0.845 & 0.795 \\
				\midrule
				\multirow[c]{4}{*}{NU}        & LR      & 0.790                   & 0.833                           & 0.751                            & 0.658 & 0.719 & 0.709 & 0.746 & 0.841 & 0.701 \\
				                              & LGBM    & 0.788                   & 0.851                           & 0.700                            & 0.704 & 0.736 & 0.732 & 0.766 & 0.857 & 0.704 \\
				                              & GEM     & 0.783                   & 0.841                           & 0.779                            & 0.731 & 0.732 & 0.776 & 0.743 & 0.846 & 0.775 \\
				                              & GEM-$*$ & 0.808                   & 0.848                           & 0.807                            & 0.738 & 0.747 & 0.785 & 0.775 & 0.859 & 0.805 \\
				\midrule
				\multirow[c]{4}{*}{MIMIC}     & LR      & 0.688                   & 0.726                           & 0.853                            & 0.638 & 0.653 & 0.779 & 0.649 & 0.757 & 0.819 \\
				                              & LGBM    & 0.726                   & 0.762                           & 0.869                            & 0.692 & 0.702 & 0.797 & 0.715 & 0.792 & 0.840 \\
				                              & GEM     & 0.768                   & 0.823                           & 0.808                            & 0.724 & 0.718 & 0.786 & 0.719 & 0.821 & 0.813 \\
				                              & GEM-$*$ & 0.793                   & 0.832                           & 0.837                            & 0.741 & 0.733 & 0.802 & 0.770 & 0.836 & 0.845 \\
				\midrule
				\multirow[c]{3}{*}{federated} & FedAvg  & 0.793                   & 0.840                           & 0.802                            & 0.737 & 0.728 & 0.780 & 0.745 & 0.837 & 0.807 \\
				                              & FedAvgM & 0.797                   & 0.838                           & 0.801                            & 0.736 & 0.730 & 0.782 & 0.748 & 0.836 & 0.804 \\
				                              & FedAdam & 0.594                   & 0.744                           & 0.659                            & 0.656 & 0.652 & 0.671 & 0.600 & 0.770 & 0.664 \\
				\midrule
				\multirow[c]{4}{*}{all}       & LR      & 0.803                   & 0.834                           & 0.831                            & 0.712 & 0.735 & 0.782 & 0.776 & 0.845 & 0.798 \\
				                              & LGBM    & 0.840                   & 0.854                           & 0.850                            & 0.739 & 0.739 & 0.793 & 0.808 & 0.860 & 0.819 \\
				                              & GEM     & 0.812                   & 0.845                           & 0.812                            & 0.746 & 0.742 & 0.793 & 0.780 & 0.848 & 0.821 \\
				                              & GEM-$*$ & 0.833                   & 0.858                           & 0.836                            & 0.753 & 0.751 & 0.806 & 0.801 & 0.864 & 0.845 \\
				\bottomrule
			\end{tabular}
		\end{tabular}
	}
	\label{tbl:roc-tkwz1}
\end{table}

%%%%%%

\begin{table}[tbh]
	\renewcommand{\tablefont}{\scriptsize}
	\tbl{\textbf{PR-AUC performance by outcome (first six)}: we report PR-AUC for each of the first six outcomes (anemia, CRRT, expired, hyperkalemia, hypernatremia, and hypertension).}{
		\centering
		\begin{tabular}{@{}c@{}}
			\begin{tabular}{llrrrrrrrrr}
				\toprule
				                              &         & \multicolumn{3}{c}{anemia} & \multicolumn{3}{c}{CRRT} & \multicolumn{3}{c}{expired}                                                 \\
				\cmidrule(lr){3-5} \cmidrule(lr){6-8} \cmidrule(lr){9-11}
				                              & outcome & UCMC                       & NU                       & MIMIC                       & UCMC  & NU    & MIMIC & UCMC  & NU    & MIMIC \\
				\midrule
				\multirow[c]{4}{*}{UCMC}      & LR      & 0.497                      & 0.231                    & 0.182                       & 0.232 & 0.094 & 0.193 & 0.597 & 0.378 & 0.366 \\
				                              & LGBM    & 0.565                      & 0.297                    & 0.271                       & 0.305 & 0.186 & 0.248 & 0.662 & 0.437 & 0.402 \\
				                              & GEM     & 0.454                      & 0.249                    & 0.232                       & 0.226 & 0.148 & 0.267 & 0.582 & 0.446 & 0.434 \\
				                              & GEM-$*$ & 0.536                      & 0.306                    & 0.288                       & 0.267 & 0.180 & 0.272 & 0.613 & 0.460 & 0.449 \\
				\midrule
				\multirow[c]{4}{*}{NU}        & LR      & 0.426                      & 0.329                    & 0.286                       & 0.199 & 0.212 & 0.222 & 0.526 & 0.591 & 0.410 \\
				                              & LGBM    & 0.439                      & 0.373                    & 0.285                       & 0.236 & 0.230 & 0.271 & 0.558 & 0.660 & 0.426 \\
				                              & GEM     & 0.475                      & 0.287                    & 0.258                       & 0.207 & 0.164 & 0.272 & 0.578 & 0.475 & 0.468 \\
				                              & GEM-$*$ & 0.531                      & 0.351                    & 0.323                       & 0.267 & 0.184 & 0.331 & 0.601 & 0.506 & 0.479 \\
				\midrule
				\multirow[c]{4}{*}{MIMIC}     & LR      & 0.361                      & 0.242                    & 0.332                       & 0.205 & 0.117 & 0.256 & 0.510 & 0.350 & 0.522 \\
				                              & LGBM    & 0.410                      & 0.287                    & 0.377                       & 0.209 & 0.149 & 0.359 & 0.533 & 0.482 & 0.580 \\
				                              & GEM     & 0.456                      & 0.245                    & 0.271                       & 0.206 & 0.155 & 0.340 & 0.561 & 0.440 & 0.463 \\
				                              & GEM-$*$ & 0.478                      & 0.280                    & 0.358                       & 0.248 & 0.166 & 0.391 & 0.626 & 0.465 & 0.538 \\
				\midrule
				\multirow[c]{3}{*}{federated} & FedAvg  & 0.457                      & 0.273                    & 0.238                       & 0.209 & 0.162 & 0.266 & 0.567 & 0.465 & 0.466 \\
				                              & FedAvgM & 0.478                      & 0.286                    & 0.258                       & 0.209 & 0.169 & 0.289 & 0.563 & 0.470 & 0.473 \\
				                              & FedAdam & 0.287                      & 0.151                    & 0.138                       & 0.063 & 0.044 & 0.082 & 0.223 & 0.171 & 0.159 \\
				\midrule
				\multirow[c]{4}{*}{all}       & LR      & 0.521                      & 0.336                    & 0.364                       & 0.289 & 0.236 & 0.306 & 0.626 & 0.590 & 0.530 \\
				                              & LGBM    & 0.543                      & 0.376                    & 0.388                       & 0.322 & 0.252 & 0.366 & 0.662 & 0.659 & 0.585 \\
				                              & GEM     & 0.516                      & 0.329                    & 0.319                       & 0.261 & 0.169 & 0.349 & 0.610 & 0.478 & 0.511 \\
				                              & GEM-$*$ & 0.545                      & 0.352                    & 0.375                       & 0.316 & 0.230 & 0.392 & 0.651 & 0.521 & 0.537 \\
				\bottomrule                                                                                                                                                                   \\
			\end{tabular}                     \\ \\
			\begin{tabular}{llrrrrrrrrr}
				\toprule
				                              &         & \multicolumn{3}{c}{hyperkalemia} & \multicolumn{3}{c}{hypernatremia} & \multicolumn{3}{c}{hypertension}                                                 \\
				\cmidrule(lr){3-5} \cmidrule(lr){6-8} \cmidrule(lr){9-11}
				                              & outcome & UCMC                             & NU                                & MIMIC                            & UCMC  & NU    & MIMIC & UCMC  & NU    & MIMIC \\
				\midrule
				\multirow[c]{4}{*}{UCMC}      & LR      & 0.049                            & 0.026                             & 0.056                            & 0.063 & 0.032 & 0.038 & 0.524 & 0.326 & 0.266 \\
				                              & LGBM    & 0.047                            & 0.035                             & 0.038                            & 0.129 & 0.047 & 0.044 & 0.560 & 0.380 & 0.335 \\
				                              & GEM     & 0.074                            & 0.053                             & 0.066                            & 0.129 & 0.098 & 0.049 & 0.558 & 0.431 & 0.422 \\
				                              & GEM-$*$ & 0.080                            & 0.064                             & 0.080                            & 0.162 & 0.118 & 0.072 & 0.577 & 0.449 & 0.434 \\
				\midrule
				\multirow[c]{4}{*}{NU}        & LR      & 0.038                            & 0.055                             & 0.050                            & 0.056 & 0.121 & 0.055 & 0.495 & 0.436 & 0.342 \\
				                              & LGBM    & 0.040                            & 0.028                             & 0.046                            & 0.073 & 0.158 & 0.027 & 0.527 & 0.458 & 0.347 \\
				                              & GEM     & 0.092                            & 0.059                             & 0.076                            & 0.142 & 0.188 & 0.084 & 0.566 & 0.452 & 0.434 \\
				                              & GEM-$*$ & 0.080                            & 0.066                             & 0.083                            & 0.153 & 0.180 & 0.063 & 0.572 & 0.466 & 0.452 \\
				\midrule
				\multirow[c]{4}{*}{MIMIC}     & LR      & 0.031                            & 0.022                             & 0.067                            & 0.042 & 0.015 & 0.073 & 0.406 & 0.367 & 0.445 \\
				                              & LGBM    & 0.034                            & 0.026                             & 0.063                            & 0.053 & 0.025 & 0.094 & 0.448 & 0.387 & 0.466 \\
				                              & GEM     & 0.070                            & 0.045                             & 0.075                            & 0.091 & 0.101 & 0.108 & 0.538 & 0.429 & 0.455 \\
				                              & GEM-$*$ & 0.069                            & 0.047                             & 0.076                            & 0.142 & 0.127 & 0.128 & 0.559 & 0.438 & 0.467 \\
				\midrule
				\multirow[c]{3}{*}{federated} & FedAvg  & 0.083                            & 0.060                             & 0.068                            & 0.163 & 0.140 & 0.078 & 0.569 & 0.432 & 0.439 \\
				                              & FedAvgM & 0.071                            & 0.058                             & 0.070                            & 0.135 & 0.166 & 0.083 & 0.562 & 0.444 & 0.445 \\
				                              & FedAdam & 0.039                            & 0.015                             & 0.042                            & 0.039 & 0.015 & 0.016 & 0.436 & 0.366 & 0.313 \\
				\midrule
				\multirow[c]{4}{*}{all}       & LR      & 0.055                            & 0.062                             & 0.082                            & 0.113 & 0.144 & 0.120 & 0.543 & 0.445 & 0.444 \\
				                              & LGBM    & 0.057                            & 0.046                             & 0.053                            & 0.152 & 0.130 & 0.093 & 0.579 & 0.465 & 0.465 \\
				                              & GEM     & 0.081                            & 0.060                             & 0.074                            & 0.188 & 0.192 & 0.160 & 0.571 & 0.453 & 0.454 \\
				                              & GEM-$*$ & 0.075                            & 0.061                             & 0.090                            & 0.223 & 0.187 & 0.153 & 0.593 & 0.467 & 0.456 \\
				\bottomrule                                                                                                                                                                                       \\
			\end{tabular} \\
		\end{tabular}
	}
	\label{tbl:pr-tkwz0}
\end{table}

\begin{table}[tbh]
	\renewcommand{\tablefont}{\scriptsize}
	\tbl{\textbf{PR-AUC performance by outcome (last six)}: we report PR-AUC for each of the last six outcomes (hypokalemia, hyponatremia, hypotension, IMV, tachycardia, and vasopressors).}{
		\centering
		\begin{tabular}{@{}c@{}}
			\begin{tabular}{llrrrrrrrrr}
				\toprule
				                              &         & \multicolumn{3}{c}{hypokalemia} & \multicolumn{3}{c}{hyponatremia} & \multicolumn{3}{c}{hypotension}                                                 \\
				\cmidrule(lr){3-5} \cmidrule(lr){6-8} \cmidrule(lr){9-11}
				                              & outcome & UCMC                            & NU                               & MIMIC                           & UCMC  & NU    & MIMIC & UCMC  & NU    & MIMIC \\
				\midrule
				\multirow[c]{4}{*}{UCMC}      & LR      & 0.043                           & 0.005                            & 0.015                           & 0.009 & 0.003 & 0.006 & 0.617 & 0.516 & 0.613 \\
				                              & LGBM    & 0.017                           & 0.006                            & 0.008                           & 0.028 & 0.008 & 0.008 & 0.673 & 0.521 & 0.579 \\
				                              & GEM     & 0.018                           & 0.011                            & 0.015                           & 0.015 & 0.022 & 0.007 & 0.651 & 0.592 & 0.738 \\
				                              & GEM-$*$ & 0.023                           & 0.011                            & 0.015                           & 0.020 & 0.028 & 0.008 & 0.675 & 0.615 & 0.761 \\
				\midrule
				\multirow[c]{4}{*}{NU}        & LR      & 0.016                           & 0.010                            & 0.029                           & 0.012 & 0.008 & 0.032 & 0.599 & 0.606 & 0.622 \\
				                              & LGBM    & 0.019                           & 0.009                            & 0.010                           & 0.005 & 0.051 & 0.007 & 0.595 & 0.620 & 0.662 \\
				                              & GEM     & 0.017                           & 0.011                            & 0.020                           & 0.013 & 0.022 & 0.007 & 0.629 & 0.626 & 0.748 \\
				                              & GEM-$*$ & 0.018                           & 0.014                            & 0.017                           & 0.007 & 0.046 & 0.009 & 0.662 & 0.641 & 0.775 \\
				\midrule
				\multirow[c]{4}{*}{MIMIC}     & LR      & 0.030                           & 0.005                            & 0.022                           & 0.003 & 0.002 & 0.007 & 0.533 & 0.496 & 0.736 \\
				                              & LGBM    & 0.014                           & 0.006                            & 0.019                           & 0.006 & 0.003 & 0.016 & 0.581 & 0.566 & 0.774 \\
				                              & GEM     & 0.021                           & 0.014                            & 0.016                           & 0.010 & 0.025 & 0.012 & 0.618 & 0.586 & 0.774 \\
				                              & GEM-$*$ & 0.026                           & 0.010                            & 0.017                           & 0.016 & 0.030 & 0.015 & 0.655 & 0.604 & 0.797 \\
				\midrule
				\multirow[c]{3}{*}{federated} & FedAvg  & 0.017                           & 0.012                            & 0.018                           & 0.015 & 0.050 & 0.011 & 0.649 & 0.616 & 0.761 \\
				                              & FedAvgM & 0.018                           & 0.013                            & 0.022                           & 0.015 & 0.036 & 0.011 & 0.654 & 0.620 & 0.761 \\
				                              & FedAdam & 0.014                           & 0.007                            & 0.010                           & 0.501 & 0.501 & 0.422 & 0.476 & 0.531 & 0.643 \\
				\midrule
				\multirow[c]{4}{*}{all}       & LR      & 0.022                           & 0.008                            & 0.026                           & 0.005 & 0.014 & 0.013 & 0.625 & 0.610 & 0.727 \\
				                              & LGBM    & 0.042                           & 0.012                            & 0.017                           & 0.009 & 0.080 & 0.010 & 0.692 & 0.621 & 0.769 \\
				                              & GEM     & 0.020                           & 0.013                            & 0.019                           & 0.023 & 0.024 & 0.013 & 0.666 & 0.637 & 0.777 \\
				                              & GEM-$*$ & 0.032                           & 0.016                            & 0.038                           & 0.020 & 0.032 & 0.014 & 0.689 & 0.647 & 0.797 \\
				\bottomrule                                                                                                                                                                                    \\
			\end{tabular} \\ \\

			\begin{tabular}{llrrrrrrrrr}
				\toprule
				                              &         & \multicolumn{3}{c}{IMV} & \multicolumn{3}{c}{tachycardia} & \multicolumn{3}{c}{vasopressors}                                                 \\
				\cmidrule(lr){3-5} \cmidrule(lr){6-8} \cmidrule(lr){9-11}
				                              & outcome & UCMC                    & NU                              & MIMIC                            & UCMC  & NU    & MIMIC & UCMC  & NU    & MIMIC \\
				\midrule
				\multirow[c]{4}{*}{UCMC}      & LR      & 0.387                   & 0.188                           & 0.200                            & 0.358 & 0.245 & 0.217 & 0.477 & 0.304 & 0.178 \\
				                              & LGBM    & 0.498                   & 0.228                           & 0.261                            & 0.386 & 0.239 & 0.270 & 0.527 & 0.369 & 0.180 \\
				                              & GEM     & 0.412                   & 0.353                           & 0.445                            & 0.397 & 0.289 & 0.342 & 0.476 & 0.586 & 0.349 \\
				                              & GEM-$*$ & 0.465                   & 0.371                           & 0.440                            & 0.426 & 0.294 & 0.336 & 0.517 & 0.590 & 0.350 \\
				\midrule
				\multirow[c]{4}{*}{NU}        & LR      & 0.374                   & 0.325                           & 0.309                            & 0.328 & 0.277 & 0.281 & 0.463 & 0.552 & 0.226 \\
				                              & LGBM    & 0.352                   & 0.348                           & 0.267                            & 0.355 & 0.293 & 0.287 & 0.495 & 0.605 & 0.225 \\
				                              & GEM     & 0.400                   & 0.361                           & 0.448                            & 0.394 & 0.296 & 0.344 & 0.487 & 0.592 & 0.348 \\
				                              & GEM-$*$ & 0.444                   & 0.382                           & 0.471                            & 0.418 & 0.309 & 0.355 & 0.514 & 0.608 & 0.364 \\
				\midrule
				\multirow[c]{4}{*}{MIMIC}     & LR      & 0.269                   & 0.192                           & 0.520                            & 0.292 & 0.227 & 0.336 & 0.392 & 0.502 & 0.362 \\
				                              & LGBM    & 0.323                   & 0.277                           & 0.572                            & 0.334 & 0.258 & 0.369 & 0.430 & 0.557 & 0.432 \\
				                              & GEM     & 0.358                   & 0.333                           & 0.479                            & 0.396 & 0.277 & 0.353 & 0.461 & 0.567 & 0.388 \\
				                              & GEM-$*$ & 0.409                   & 0.359                           & 0.520                            & 0.407 & 0.295 & 0.368 & 0.517 & 0.586 & 0.435 \\
				\midrule
				\multirow[c]{3}{*}{federated} & FedAvg  & 0.408                   & 0.366                           & 0.458                            & 0.413 & 0.293 & 0.343 & 0.472 & 0.600 & 0.365 \\

				                              & FedAvgM & 0.420                   & 0.359                           & 0.460                            & 0.408 & 0.299 & 0.347 & 0.477 & 0.582 & 0.373 \\

				                              & FedAdam & 0.207                   & 0.280                           & 0.247                            & 0.303 & 0.232 & 0.232 & 0.313 & 0.555 & 0.198 \\
				\midrule
				\multirow[c]{4}{*}{all}       & LR      & 0.424                   & 0.335                           & 0.481                            & 0.373 & 0.292 & 0.354 & 0.497 & 0.584 & 0.356 \\
				                              & LGBM    & 0.522                   & 0.362                           & 0.549                            & 0.402 & 0.294 & 0.361 & 0.543 & 0.607 & 0.399 \\
				                              & GEM     & 0.447                   & 0.369                           & 0.490                            & 0.417 & 0.298 & 0.369 & 0.526 & 0.590 & 0.385 \\
				                              & GEM-$*$ & 0.496                   & 0.409                           & 0.531                            & 0.439 & 0.309 & 0.389 & 0.553 & 0.616 & 0.428 \\
				\bottomrule
			\end{tabular}
		\end{tabular}
	}
	\label{tbl:pr-tkwz1}
\end{table}

\begin{figure}[tbh]
	\centering
\centering
		\includegraphics[width=\textwidth]{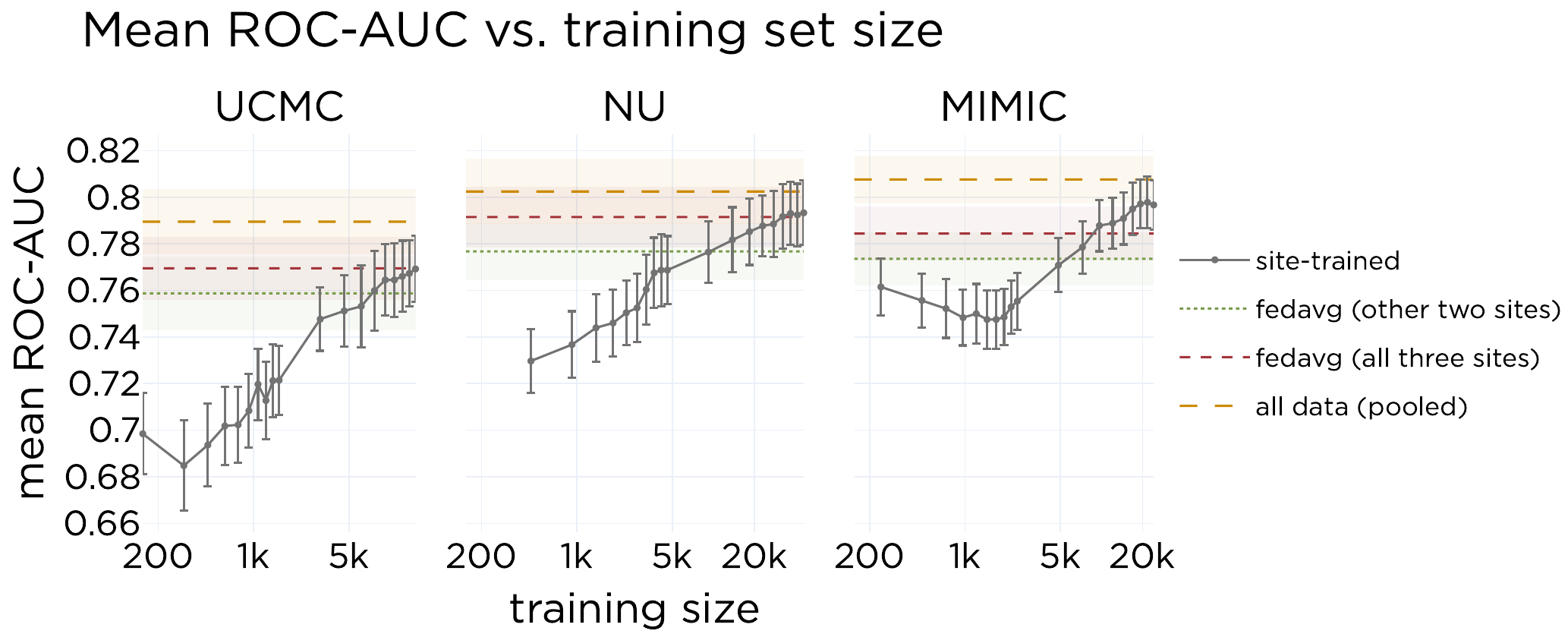}
		\caption{PR-AUC performance on each respective fixed test set vs. number of training examples available. NU is the largest dataset and so has the widest graph. Error bars indicate 95\% confidence intervals for performance. Horizontal lines represent (in ascending order) performance of a federated model trained on the \emph{other} two sites, performance of a federated model trained on all sites, and performance of a model trained on a centralized aggregate dataset. See Figure~\ref{fig:local-roc-auc} for ROC-AUC.}
		\label{fig:local-pr-auc}
\end{figure}

\clearpage
\appendix{Site Heterogeneity Analysis}

\begin{figure}[tbh!]
\centering
    \includegraphics[width=\textwidth]{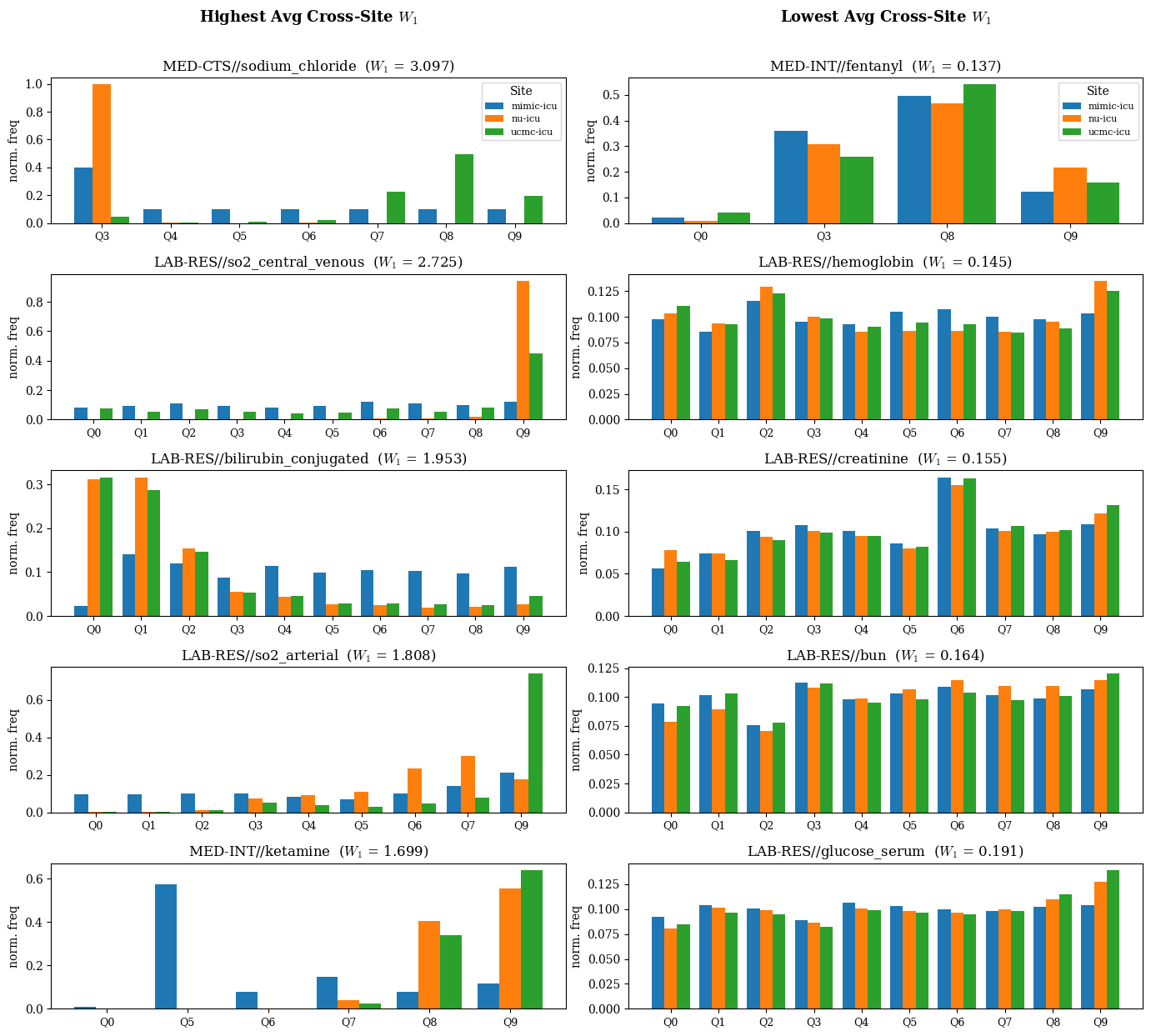}
    \caption{Distributions of the quantile bins across each of the data sites. Recall that decile bins are initialized according to the MIMIC training split. Missing bins, such as in the case of MED-INT//fentanyl Q1 and Q2, occur when adjacent deciles are equal, with the highest of the equal deciles being used to tokenize all entries within the bin. \textbf{Left:} The bin distributions for the value categories where the mean Wasserstein-1 distance between sites is the largest. These are the value categories where the observed distribution of quantile tokens varies the most between sites. \textbf{Right:} The bin distributions for the value categories where the mean Wasserstein-1 distance between sites is the smallest. These are the value categories where the observed distribution of quantile tokens remains the most consistent between sites.}
    \label{fig:cross-site-bins}
\end{figure}
\newpage
\begin{figure}[tbh]
	\centering
    \includegraphics[width=0.7\textwidth]{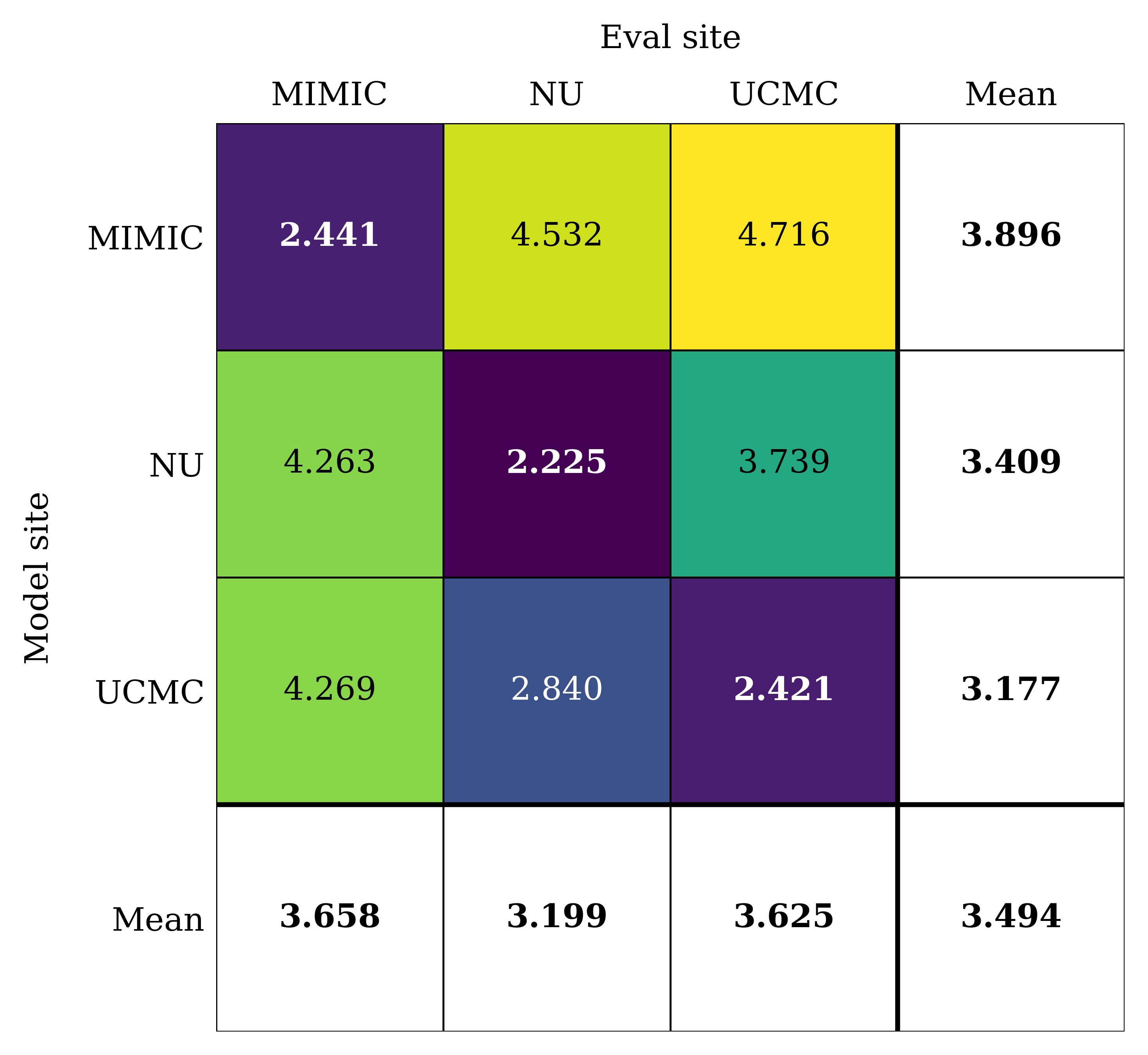}
    \caption{Cross-site evaluation of contextual next-token prediction surprise~\cite{Bur26} (or NLL) for all combinations of training site and evaluation site. Average surprise is computed by taking the average of the negative log of the predicted probability of all observed tokens at each token in a pool of timelines. For this experiment, a pool of $5,000$ timelines are sampled from each site and used to compute the average surprise. As expected, the model trained at each site is least surprised by data the same site. MIMIC, which contains data recorded on the other side of the country multiple years prior to the other data sites, trains a model that generates the highest mean surprise across all sites. The model trained on UCMC data, on the other hand, experiences the least surprise on average. However, the UCMC samples generate mean surprise only marginally lower than the MIMIC samples.}
    \label{fig:cross-site-entropy}
\end{figure}
\end{document}